\documentclass[10pt,journal,compsoc]{IEEEtran}
\usepackage[colorlinks, linkcolor=red, urlcolor=blue,
anchorcolor=red, citecolor=green]{hyperref}
\usepackage{graphicx}
\usepackage{amsmath,amsfonts}
\usepackage{algorithmic}
\usepackage{algorithm}
\usepackage{array}
\usepackage{textcomp}
\usepackage{stfloats}
\usepackage{url}
\usepackage{verbatim}
\usepackage{graphicx}
\usepackage{cite}
\usepackage{booktabs} 
\usepackage{multirow}

\ifCLASSINFOpdf
\else
\fi
\begin{document}
\title{MedUniSeg: 2D and 3D Medical Image Segmentation via a Prompt-driven Universal Model}

\author{Yiwen Ye,
Ziyang Chen,
Jianpeng Zhang,
Yutong Xie,
and
Yong Xia,~\IEEEmembership{Member,~IEEE}

\IEEEcompsocitemizethanks{\IEEEcompsocthanksitem This work was supported in part by the National Natural Science Foundation of China under Grants 62171377. ({\em Corresponding authors: Y. Xie and Y. Xia}).
\IEEEcompsocthanksitem Y. Ye and Z. Chen are with the National Engineering Laboratory for Integrated Aero-Space-Ground-Ocean Big Data Application Technology, School of Computer Science and Engineering, Northwestern Polytechnical University, Xi’an 710072, China. \protect\\
E-mail: \{ywye, zychen\}@mail.nwpu.edu.cn.
\IEEEcompsocthanksitem J. Zhang is with the College of Computer Science and Technology, Zhejiang University, Zhejiang, China. \protect\\
E-mail: jianpeng.zhang0@gmail.com.
\IEEEcompsocthanksitem Y. Xie is with the Australian Institute for Machine Learning (AIML), The University of Adelaide, Australia. \protect\\
E-mail: yutong.xie678@gmail.com.
\IEEEcompsocthanksitem Y. Xia is with the National Engineering Laboratory for Integrated Aero-Space-Ground-Ocean Big Data Application Technology, School of Computer Science and Engineering, Northwestern Polytechnical University, Xi’an 710072, China, with Research \& Development Institute of Northwestern Polytechnical University in Shenzhen, Shenzhen 518057, China, and also with the Ningbo Institute of Northwestern Polytechnical University, Ningbo 315048, China. \protect\\ 
E-mail: yxia@nwpu.edu.cn.
}
}% <-this % stops an unwanted space
% \thanks{Manuscript received April 19, 2005; revised August 26, 2015.}}


\markboth{Journal of \LaTeX\ Class Files,~Vol.~14, No.~8, August~2015}%
{Shell \MakeLowercase{\textit{et al.}}: Bare Demo of IEEEtran.cls for Computer Society Journals}



\IEEEtitleabstractindextext{%
\begin{abstract}
Universal segmentation models offer significant potential in addressing a wide range of tasks by effectively leveraging discrete annotations. As the scope of tasks and modalities expands, it becomes increasingly important to generate and strategically position task- and modal-specific priors within the universal model. However, existing universal models often overlook the correlations between different priors, and the optimal placement and frequency of these priors remain underexplored. 
In this paper, we introduce MedUniSeg, a prompt-driven universal segmentation model designed for 2D and 3D multi-task segmentation across diverse modalities and domains. MedUniSeg employs multiple modal-specific prompts alongside a universal task prompt to accurately characterize the modalities and tasks. To generate the related priors, we propose the modal map (MMap) and the fusion and selection (FUSE) modules, which transform modal and task prompts into corresponding priors. These modal and task priors are systematically introduced at the start and end of the encoding process. 
We evaluate MedUniSeg on a comprehensive multi-modal upstream dataset consisting of 17 sub-datasets. The results demonstrate that MedUniSeg achieves superior multi-task segmentation performance, attaining a 1.2\% improvement in the mean Dice score across the 17 upstream tasks compared to nnUNet baselines, while using less than $1/10$ of the parameters. For tasks that underperform during the initial multi-task joint training, we freeze MedUniSeg and introduce new modules to re-learn these tasks. This approach yields an enhanced version, MedUniSeg*, which consistently outperforms MedUniSeg across all tasks. Moreover, MedUniSeg surpasses advanced self-supervised and supervised pre-trained models on six downstream tasks, establishing itself as a high-quality, highly generalizable pre-trained segmentation model.
The code and model will be available at \href{https://github.com/yeerwen/UniSeg}{https://github.com/yeerwen/UniSeg}.
\end{abstract}

\begin{IEEEkeywords}
 Medical image segmentation, Universal model, Prompt learning, Multi-modal learning
\end{IEEEkeywords}}

\maketitle

\IEEEdisplaynontitleabstractindextext

% \fi
\IEEEpeerreviewmaketitle

\begin{figure}[t]
  \centering
  \includegraphics[width=0.95\linewidth]{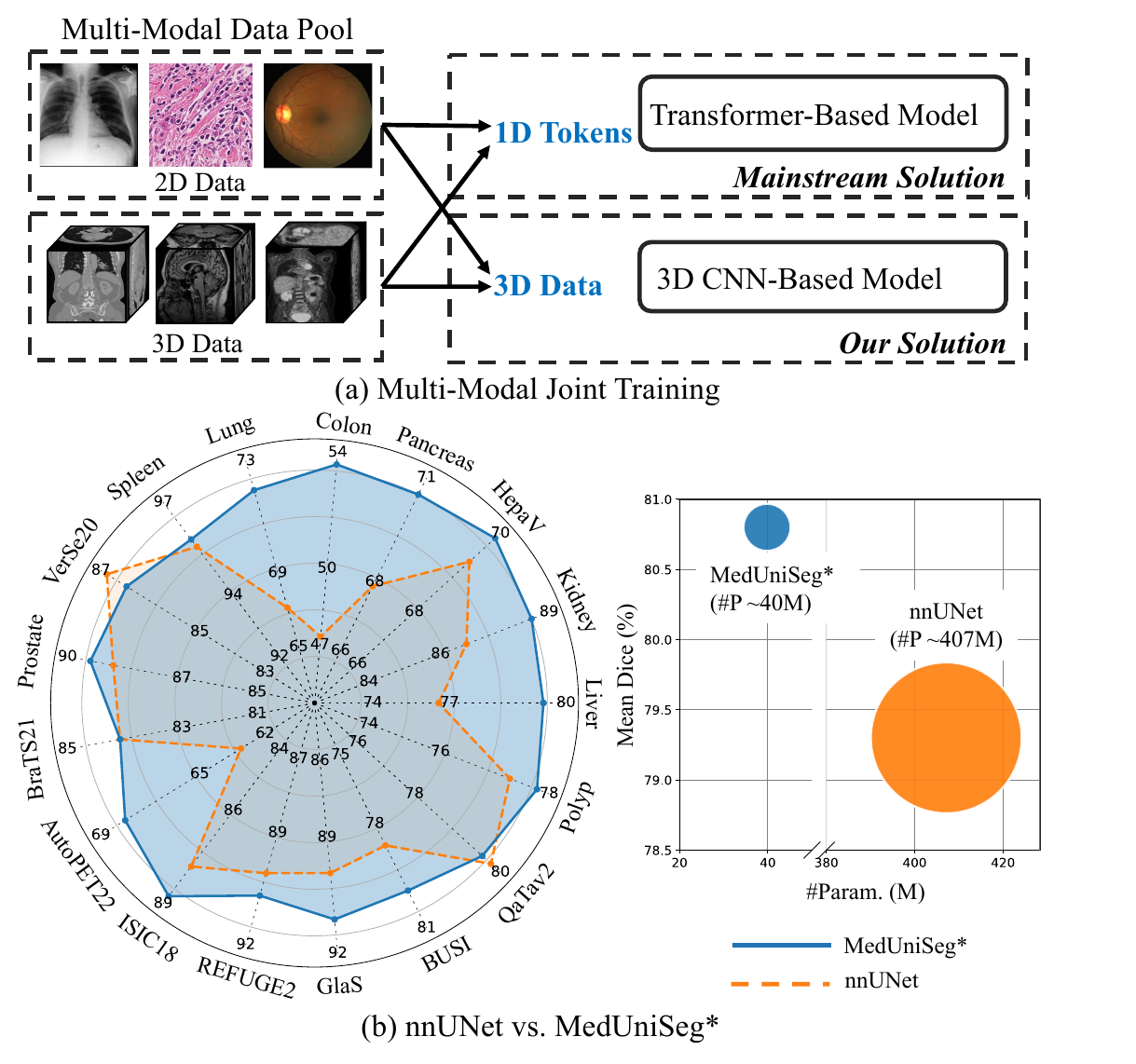}
   \caption{
   (a) Comparison between the mainstream solution and our solution. The mainstream solution treats both 2D and 3D data as 1D tokens and utilizes a Transformer-based model for processing. In contrast, our solution interprets 2D data as pseudo-3D data and employs a 3D CNN-based model for processing.
   (b) Performance and parameter comparisons between nnUNet and MedUniSeg* across 17 upstream datasets. To achieve the same tasks, nnUNet requires 17 individual models, comprising 11 3D models and 6 2D models, while our MedUniSeg* needs only a single model.
   }
\label{fig: intro}
\end{figure}

\IEEEraisesectionheading{\section{Introduction}\label{sec:introduction}}

\IEEEPARstart{M}{edical} image segmentation is essential for delineating lesions, diagnosing diseases, analyzing pathology, and planning treatments. With the diversification of imaging techniques and targets, many segmentation tasks now involve various data modalities and anatomical regions, covering both 2D and 3D data. The advent of deep learning has facilitated automated methods to address these tasks effectively. However, two main challenges remain: (1) the tendency to create specialized models for specific tasks, which leads to fragmented research efforts, and (2) the limitation of small labeled datasets, particularly for 3D segmentation, due to the labor-intensive nature of voxel-wise annotations.

Universal models that can tackle multiple segmentation tasks through a single training process have emerged as a promising solution. These models utilize extensive data from various datasets to enhance learning. A key aspect of their design is determining the task-related priors to incorporate and their optimal placement in the model for effective task awareness. One intuitive approach employs a shared encoder with multiple task-specific decoders \cite{chen2019med3d}, but this can result in structural redundancy and parameter inefficiency due to the multiple branches needed, especially when integrating numerous tasks. To streamline the model structure, some universal models transform multi-dataset training into multi-class training by assigning each target a unique output channel \cite{ulrich2023multitalent, zhou2019prior, fang2020multi, huang2020multi, liu2022universal, zhang2023merging, liu2024cosst, chen2024versatile, liu2023clip, liu2024universal}. These models derive task-related priors by selecting the corresponding segmentation head for each task. Additionally, some prompt-based universal models utilize fixed task-specific prompts \cite{zhang2021dodnet, wu2022tgnet, deng2023omni}, such as one-hot encoding, or learnable task-specific prompts \cite{xie2023learning}, to introduce task-related priors at the end of the decoder stage. These models, however, often struggle in complex and varied segmentation scenarios, as only a few parameters are aware of the current task; thus, task-related priors are integrated too late in the process. In our previous work, UniSeg \cite{ye2023uniseg}, we addressed this challenge by adding task-related prior to the end of the encoding process, enabling the whole decoder to be aware of tasks.
Recently, models like CCQ \cite{liu2023ccq} and Hermes \cite{gao2024training} have sought to enhance task-related information by introducing learnable prompts at multiple stages throughout the model. Despite these advancements, the relationships between different tasks remain less explored, and the optimal locations and frequencies for introducing these priors require further refinement.

Moreover, current universal segmentation methods primarily focus on either single-modal segmentation \cite{dmitriev2019learning, zhang2021dodnet, deng2023omni, wu2022tgnet, xie2023learning, liu2023ccq} or single-dimensional segmentation \cite{dmitriev2019learning, zhang2021dodnet, deng2023omni, wu2022tgnet, xie2023learning, gao2024training, liu2023ccq}, failing to meet the multi-modal and multi-dimensional requirements of medical image segmentation. Therefore, developing a generalized universal model capable of processing multi-modal and multi-dimensional data is essential. Constructing such a model faces two primary challenges: first, a backbone is needed that delivers superior segmentation performance while accommodating inputs of varying dimensions, including both 2D and 3D data. Second, the significant differences between modalities pose a risk of optimization conflicts during joint training \cite{ye2024continual, huang2022modality, wang2020makes}.

%Furthermore, current universal segmentation methods primarily focus on either single-modal segmentation \cite{dmitriev2019learning,zhang2021dodnet,deng2023omni,wu2022tgnet,xie2023learning,liu2023ccq} or single-dimensional segmentation \cite{dmitriev2019learning,zhang2021dodnet,deng2023omni,wu2022tgnet,xie2023learning,gao2024training,liu2023ccq}, which does not meet the multi-modal and multi-dimensional requirements of medical image segmentation. 
%Therefore, it is crucial to develop a more generalized universal model capable of processing multi-modality and multi-dimensional data. 
% Therefore, it is crucial to develop a more generalized universal model capable of processing a wide range of medical data types, including multi-modal data (\textit{e.g.}, CT scans, MR scans, and X-ray images) and multi-dimensional data (\textit{e.g.}, 2D data and 3D data). 
%The construction of such a model faces two primary challenges. First, there is a need for a backbone that not only delivers superior segmentation performance but also accommodates inputs of varying dimensions, including 2D and 3D data. Second, the significant differences between modalities pose a risk of optimization conflicts during joint training \cite{ye2024continual,huang2022modality,wang2020makes}. 

To address these limitations, we propose a prompt-driven \textbf{Med}ical \textbf{Uni}versal \textbf{Seg}mentation model (MedUniSeg). This model is designed to segment multiple organs, tissues, vertebrae, tumors, and lesions in 2D and 3D medical images across various modalities and domains. The architecture of MedUniSeg comprises several components: a modal map (MMap) module, a vision encoder, a fusion and selection (FUSE) module, and a prompt-driven decoder. The MMap and FUSE modules leverage prompt learning to provide modal-specific and task-specific priors, respectively, thereby alleviating optimization conflicts between modalities and enhancing task-related progress. Specifically, the MMap module maps learnable modal-specific prompts to align with the shape of the input image, enriching the input data with modal-specific priors. The FUSE module integrates a learnable universal task prompt, which describes the correlations between tasks, and the features from the vision encoder to generate task-specific priors. We employ multiple modal-specific prompts and a universal task prompt based on the premise that \textit{potential correlations exist between different tasks, while correlations among modalities are negligible, primarily due to the use of unpaired multi-modal data \cite{godau2021task}.} 
Furthermore, we carefully consider the integration locations for modal-specific and task-specific priors. Modal-specific priors are introduced at the start of the encoding process to guide different modality data, while task-specific priors are introduced at the end of the encoding process to meet the specific needs of distinct segmentation tasks. \textit{The differing locations depend on when discrepancies between modalities or tasks begin to emerge.} Since different modalities necessitate distinct feature extraction procedures, the model must address these variations early in the encoding process. After extracting high-level semantic features, different tasks correspond to specific decoding processes; thus, the model must be informed of the task priors at the onset of this stage.

To effectively handle most segmentation tasks, our model must accept both 2D and 3D input data. Unlike the prevailing trend of using Transformer-based models that process data in a sequence-to-sequence manner, MedUniSeg adopts a novel perspective by treating 2D data as pseudo-3D data with a depth of one and employing a pruned 3D CNN-based UNet to manage both 2D and 3D data (see Fig. \ref{fig: intro}(a)). Although this approach demands more resources than its 2D-only counterparts for predicting 2D segmentation maps, MedUniSeg still surpasses Transformer-based models like UniMiSS in terms of inference time and performance (see Section \ref{inference}).

%To handle most segmentation tasks, our model must accept both 2D and 3D input data. Unlike the prevailing trend of using Transformer-based models, which handle data in a sequence-to-sequence manner, MedUniSeg provides a new perspective, treating 2D data as pseudo 3D data with a depth of one and employing a pruned 3D CNN-based UNet to address both 2D and 3D data, as shown in Fig. \ref{fig: intro}(a). Although requiring more resources than its 2D version models for predicting 2D segmentation maps, MedUniSeg still outperforms Transformer-based models like UniMiSS in terms of inference time and performance (see Section \ref{inference}).

% To handle most segmentation tasks, our model must accept both 2D and 3D input data. Unlike the prevailing trend of using Transformer-based models, which handle data in a sequence-to-sequence manner, MedUniSeg employs a pruned 3D CNN-based UNet for both 2D and 3D data, as shown in Fig. \ref{fig: intro}(a). For 2D data, we treat them as pseudo 3D data with a depth of one. This decision is based on three critical observations: (1) Most medical segmentation tasks involve 2D or 3D images; (2) CNN-based backbones like nnUNet have demonstrated superior performance over Transformer-based counterparts in segmentation tasks, particularly in 3D segmentation tasks (see Section \ref{exp: upstream}); (3) although requiring more resources than its 2D version models for predicting 2D segmentation maps, MedUniSeg still outperforms Transformer-based models like UniMiSS in terms of inference time and performance (see Section \ref{inference}).
%

For evaluation, we compiled a comprehensive dataset comprising 21,382 3D/2D samples across nine modalities (CT, MRI, PET, dermoscopy, fundus imaging, pathological imaging, ultrasound, X-ray, and endoscopy) and 24 targets from 17 datasets, referred to as upstream datasets. We benchmarked MedUniSeg against other universal models like DoDNet \cite{zhang2021dodnet} and Hermes \cite{gao2024training}, as well as leading single-task models like nnUNet \cite{isensee2021nnu}, U-Mamba \cite{ma2024u}, and UKAN \cite{li2024u}, each trained independently on their respective datasets. The results demonstrate that MedUniSeg achieves superior generalization performance across all upstream tasks, with only a few tasks slightly underperforming compared to nnUNet models, which serve as our baselines. To further enhance performance, we froze the trained model and integrated new LoRA \cite{hu2021lora}, deconvolutional layers, and segmentation heads to re-learn these tasks, resulting in an enhanced version, MedUniSeg*. Performance and parameter comparisons between nnUNet and MedUniSeg* are illustrated in Fig. \ref{fig: intro}(b). The visualization indicates that MedUniSeg* outperforms nnUNet on 14 tasks, with only marginally lower performance on two tasks, while utilizing less than $1/10$ of the parameters. To assess the transfer capability of MedUniSeg, we fine-tuned it on six downstream datasets and conducted comparative analyses against other universal models and self-supervised models such as VoCo \cite{wu2024voco} and MedKLIP \cite{wu2023medklip}. The results reveal that MedUniSeg outperforms all competitors regarding generalization performance across the 17 upstream tasks and six downstream tasks.

The contributions of this work are four-fold:
\begin{itemize}
    \item We further explore the universal medical segmentation model, enhancing its capability across different modalities and data dimensions. Our model can simultaneously address 17 segmentation tasks spanning nine modalities, various domains, and both 2D and 3D dimensions, using a single model built upon UNet.
    \item We design two types of learnable prompts to generate specific priors tailored to the modality and task of the ongoing image. Additionally, we customize the introduction locations of the proposed priors to mitigate modal collisions and facilitate task learning. 
    \item We introduce LoRA to improve the performance of tasks that do not benefit from joint training, thereby contributing to a more comprehensive and versatile universal segmentation model.
    \item MedUniSeg serves as a high-performance pre-trained model for both 2D and 3D medical image segmentation, demonstrating strong generalization and high-quality representation capabilities.
\end{itemize}

%=====================================================================================================================================
%===========================================================Related Work==============================================================
%=====================================================================================================================================

\section{Related Work}
\subsection{Universal Model for Medical Image Segmentation}
The diverse modalities in medical imaging, coupled with labor-intensive annotation processes and disease-specific variations, often lead to fragmented annotation efforts across multiple segmentation datasets. Traditionally, each dataset is managed by a separate model, which results in distributed research efforts. To counter this fragmentation, the development of universal models capable of handling multiple datasets or tasks has gained traction and shown considerable promise. These universal models are typically categorized into three groups: multi-head models, multi-class models, and prompt-based models.

%Various modalities, labor-intensive annotation processes, and disease-specific variations often disperse annotation efforts across numerous segmentation datasets, each traditionally addressed by a specific model. This approach has led to distributed research efforts.
%Consequently, the development of single models capable of addressing multiple datasets, \textit{i.e.}, multiple tasks, has emerged, showing considerable promise. These universal-purpose models are typically categorized into three groups: multi-head models, multi-class models, and prompt-based models.

%
\textbf{Multi-head models} generally utilize a shared encoder combined with multiple task-specific decoders \cite{chen2019med3d}. While this architecture facilitates task integration to optimize parameter utilization, it also introduces redundancy and increases model complexity.
\textbf{Multi-class models} consolidate multiple tasks into a single multi-class task, assigning each task to a specific channel within the output segmentation maps. Techniques such as generating pseudo labels \cite{zhou2019prior,huang2020multi,liu2022universal,zhang2023merging,liu2024cosst}, self-disambiguation learning \cite{chen2024versatile}, target adaptive loss \cite{fang2020multi}, and masked back-propagation \cite{liu2023clip,liu2024universal,ulrich2023multitalent} are employed to integrate tasks and leverage their joint learning. For instance, the Universal Model \cite{liu2023clip,liu2024universal} employs a language-driven parameter generator to derive rich semantic encodings for each foreground category and incorporates a masked back-propagation strategy for improved learning from available annotations. However, task-related priors are primarily introduced at the segmentation heads, resulting in a limited number of parameters being `aware' of the ongoing task. This limitation hinders the model's ability to handle numerous segmentation tasks, especially in complex scenarios.
\textbf{Prompt-based models} leverage well-designed prompts to inform the model about the current task, thereby enhancing segmentation accuracy. Prompts can be fixed features associated with the target task \cite{dmitriev2019learning,zhang2021dodnet,deng2023omni,wu2022tgnet} or learnable task-specific features \cite{xie2023learning}. For instance, DoDNet \cite{zhang2021dodnet} utilizes one-hot encoding for each task as a prior, along with a dynamically generated convolutional block tailored to the ongoing task and image. TransDoDNet \cite{xie2023learning} employs learnable task-specific organ embeddings and a filters prediction head to produce task-specific filters for dynamic segmentation. Similar to multi-class models, prompt-based models introduce task-related prior information at the end of the decoder, which can hinder their performance in complex segmentation scenarios, especially as the number of modalities and tasks increases. Recently, CCQ \cite{liu2023ccq} developed a cross-class query learning module to generate class-relevant features for segmentation, introducing task-related priors at both the start and end of the decoding process. Hermes \cite{gao2024training} employs a context-prior pool to apply task- and modal-specific priors based on the input image, incorporating priors at multiple stages. However, despite the earlier introduction of task-related priors, the optimal locations and frequency of these priors remain to be refined.

In our pilot study \cite{ye2023uniseg}, we introduced UniSeg, a prompt-driven model that incorporates task priors at the end of the encoding process to enhance decoder performance. Nonetheless, UniSeg has notable limitations: it overlooks the risk of modal collision during multi-modal learning and is confined to 3D segmentation tasks, which restricts its applicability. To address these limitations, we propose MedUniSeg, which incorporates modal-specific prompts to generate modal priors and extends the model's capabilities to efficiently handle both 2D and 3D segmentation tasks under a single framework.

\subsection{Learning from Multi-modal Medical Data}
Multi-modal learning enables models to learn from diverse paired or unpaired multi-modal data and has garnered significant interest in the research community. A wide range of applications has been explored, such as multi-modal pre-training \cite{xie2024pairaug,liu2023improving,liu2023m,zhang2023knowledge,jin2023gene,ye2024continual,xie2022unimiss}, multi-modal segmentation \cite{zhang2021modality,yang2023flexible}, and multi-modal classification \cite{shao2020multi,tang2022fusionm4net}.
As research on multi-modal learning deepens, issues such as modal data collision or modality competition \cite{ye2024continual,huang2022modality,wang2020makes} arise due to significant gaps between modalities and inconsistent optimization strategies, hindering the performance of joint training. A multiway strategy, which assigns dedicated modules for each modality, effectively mitigates optimization inconsistency. For instance, BEiT-3 \cite{wang2023image} employs both vision and language expert modules across multiple transformer layers to capture vision- and language-specific features, respectively. However, this approach can lead to uncontrolled parameter growth as the number of modalities increases.
%As research on multi-modal learning deepens, modal data collision (or modality competition) \cite{ye2024continual,huang2022modality,wang2020makes}, which is caused by significant gaps between modalities and inconsistent optimization strategies, hinders the performance of joint training.
%A multiway strategy, which assigns a set of modules for each modality, is an effective way to mitigate optimization inconsistency. For example, BEiT-3 \cite{wang2023image} employs vision expert and language expert modules across multiple transformer layers to capture vision- and language-specific features, respectively. However, this approach suffers from uncontrolled parameter growth as the number of modalities increases.
An alternative strategy involves separating the training process for each modality. For example, MedCoSS \cite{ye2024continual} shifts from joint pre-training to a multi-stage pre-training approach, designating each stage for specific modal data. Although this method effectively mitigates catastrophic forgetting using continual-based techniques, it may still result in some degree of forgetting, yielding performance comparable to single-modal pre-training. In this study, we employ prompt learning to provide modal priors for the model, offering a novel perspective to address modal data collision.

Furthermore, it is crucial for models to handle both 2D and 3D data, as these encompass the majority of medical image segmentation tasks. Current methods primarily utilize Transformer-based architectures \cite{ye2024continual,xie2022unimiss,cai2022uni4eye,chen2024transunet}, which are favored for their ability to process data in a sequence-to-sequence manner. In this study, we propose treating 2D data as pseudo-3D by considering the depth dimension as one, allowing a 3D model to accommodate both 2D and 3D data. This unified approach simplifies the model structure while maintaining high performance.
%Moreover, it is essential for the model to handle both 2D and 3D data, as these include most medical image segmentation tasks. The prevalent methods involve using Transformer-based architectures \cite{ye2024continual,xie2022unimiss,cai2022uni4eye}, favored for their ability to process data in a sequence-to-sequence manner. 
%In this study, we provide another solution, which treats 2D data as pseudo 3D by considering the depth length as one and therefore adopts a 3D model to accept both 2D and 3D data. This unified strategy for 2D and 3D segmentation simplifies the model structure while maintaining high performance.

% However, this approach faces significant drawbacks: (1) a considerable portion of the parameters remains unshared due to reliance on dimension-sensitive convolutional neural network (CNN) modules, leading to redundancy; and (2) Transformer-based models usually do not outperform CNN-based models in segmentation tasks, particularly with 3D data.
% To maintain high segmentation performance, CNN-based models, such as nnUNet \cite{isensee2021nnu}, are more suitable for medical segmentation tasks. Therefore, we designed a 3D model that accepts 3D data and treats 2D data as pseudo 3D by considering the depth length as one. 

\begin{figure*}[t]
  \centering
  \includegraphics[width=0.95\linewidth]{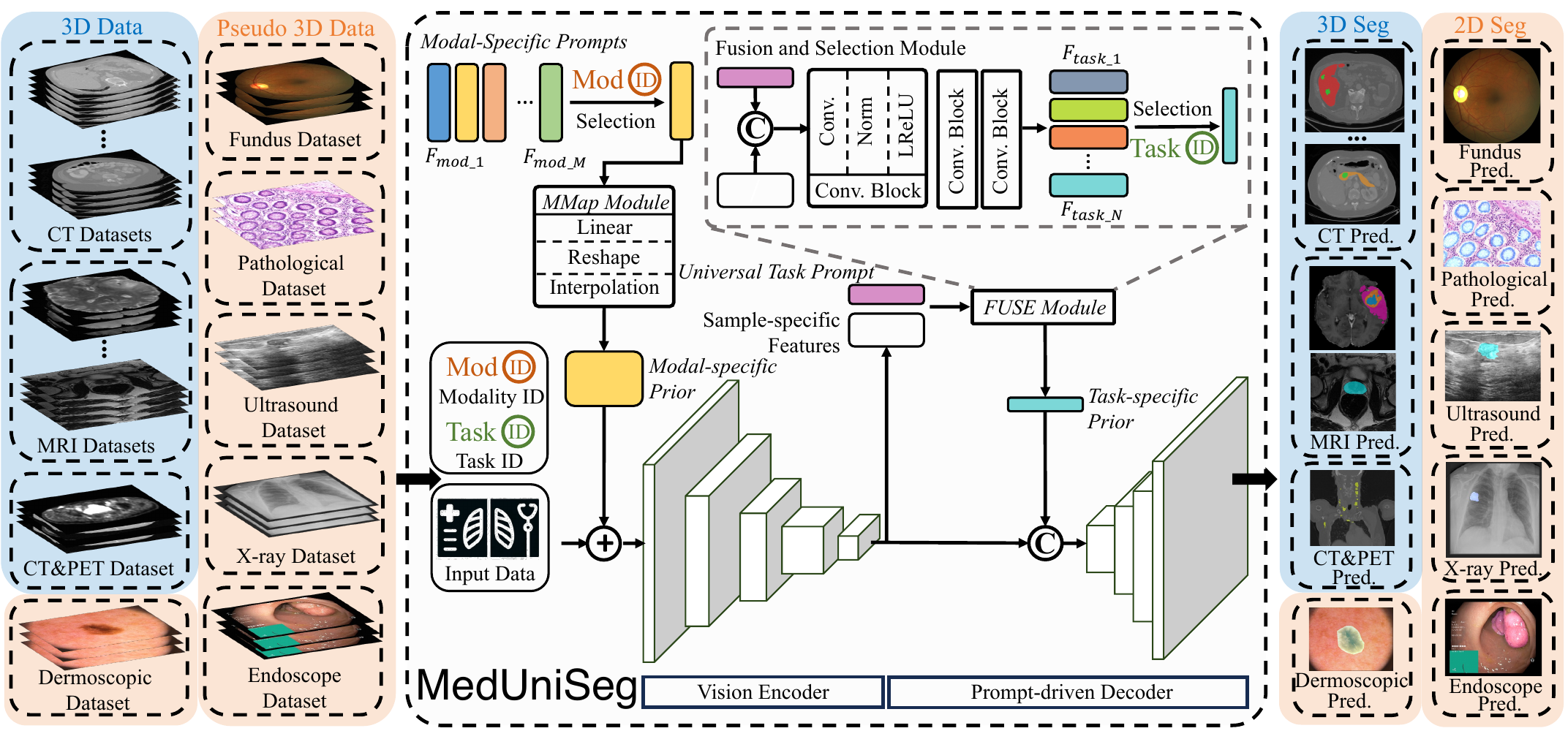}
   \caption{Technical pipeline of our MedUniSeg, including the MMap module, a vision encoder, the FUSE module, and a prompt-driven decoder. For an input image, we identify its modality ID and task ID. Based on these identifiers, the MMap module generates modal-specific priors, while the FUSE module produces task-specific priors. These priors are integrated at the start and end of the encoding process, enabling MedUniSeg to effectively handle multiple modalities and tasks.}
\label{fig: overview}
\end{figure*}

\subsection{Prompt Learning}
Prompt learning has emerged as an effective strategy for enhancing model adaptability to specific tasks by integrating prior knowledge into the model. This technique has been widely applied across various fields, including the efficient fine-tuning of large models \cite{wei2022finetuned,zhou2022conditional}, domain adaptation \cite{ge2023domain}, continual learning \cite{wang2022learning}, self-supervised learning \cite{wang2023position}, and federated learning \cite{feng2023learning}.
The effectiveness of prompt learning is particularly evident in the development of universal segmentation models, where it ensures that the model remains acutely `aware' of the current task and modality. For instance, models like DoDNet \cite{zhang2021dodnet} and its variants \cite{wu2022tgnet,deng2023omni} employ one-hot encoding as a fixed prompt. In contrast, TransDoDNet \cite{xie2023learning}, Hermes \cite{gao2024training}, and CCQ \cite{liu2023ccq} utilize learnable vectors as learnable prompts to indicate the ongoing task.
Distinct from these existing methods, this study tailors both task and modal prompts, carefully determining their introduction locations within the model's architecture. Our approach, therefore, establishes a coherent framework for multi-modal universal segmentation, significantly enhancing the model's ability to integrate and process diverse data types and tasks.

\section{Method}
\subsection{Problem Deﬁnition}
Consider the set $\{S^{1}_{1}, S^{1}_{2},...,S^{M}_{N}\}$, where $N$ datasets contain $M$ modalities. Here, $S^{m}_{i}=\{X_{ij}^{m},Y_{ij}\}^{n_i}_{j=1}$ denotes that the $i$-th dataset corresponds to the $m$-th modality and comprises $n_i$ image-annotation pairs, with $X_{ij}^{m}$ representing the image and $Y_{ij}$ the corresponding ground truth annotation. 
Traditionally, addressing these $N$ datasets necessitates training $N$ separate models, each tailored to a specific dataset. This conventional approach has significant drawbacks: (1) it disperses research efforts across multiple individual tasks, and (2) it fails to utilize the rich and diverse information available across different datasets. To overcome these limitations, we propose MedUniSeg, a universal segmentation model designed to manage multiple tasks across various modalities under a single framework. An overview of MedUniSeg is presented in Fig. \ref{fig: overview}.
%Consider $\{S^{1}_{1}, S^{1}_{2},...,S^{M}_{N}\}$, where $N$ datasets contain $M$ modalities. Here $S^{m}_{i}=\{X_{ij}^{m},Y_{ij}\}^{n_i}_{j=1}$ presents that the $i$-th dataset is $m$-th modality and has a total of $n_i$ image-label pairs, and $X_{ij}^{m}$ and $Y_{ij}$ are the image and the corresponding ground truth, respectively. Traditionally, addressing these $N$ datasets requires training $N$ separate models, each tailored to one dataset. This conventional approach has significant drawbacks: (1) it disperses research efforts across multiple individual tasks, and (2) it fails to utilize the rich and diverse information available across different datasets. To overcome both limitations, we propose a universal segmentation model, MedUniSeg, designed to handle multiple tasks across various modalities using a single model framework without any task-specific parameters. An overview of MedUniSeg is presented in Fig. \ref{fig: overview}.

\subsection{Encoder-decoder backbone}
The core architecture of MedUniSeg is based on nnUNet \cite{isensee2021nnu} and comprises a vision encoder, a decoder, and a segmentation head, all shared across different tasks. The encoder includes six stages, each featuring two convolutional blocks to extract features while progressively reducing the resolution of the feature map. Each convolutional block consists of a convolutional layer, followed by instance normalization and a LeakyReLU activation. Notably, the first convolutional layer in each stage, except the initial one, employs a stride of 2 to decrease resolution. 
%The main architecture of MedUniSeg is based on nnUNet \cite{isensee2021nnu} and contains a vision encoder, a decoder, and a segmentation head that are shared across different tasks. The encoder includes six stages, each equipped with two convolutional blocks designed to extract features while progressively reducing the resolution of the feature map. Each convolutional block consists of a convolutional layer followed by instance normalization and a LeakyReLU activation function. Notably, the first convolutional layer in each stage, except the initial one, employs a stride of 2 to decrease resolution. 
To accommodate multi-modality inputs, we modify the first convolutional layer of the model by incorporating four specific convolutional layers tailored to handle inputs with one, two, three, or four channels, respectively. The outputs from the encoder are sample-specific features, denoted as $F \in \mathbb{R}^{C_{1}\times\frac{D}{16}\times\frac{H}{32}\times\frac{W}{32}}$, where $C_{1}$ is the number of channels, and $D$, $H$, and $W$ indicate the depth, height, and width of the input, respectively.
%To accept the multi-modality inputs, the architecture includes modifications to the first convolutional layer of the model, where four specific convolution layers are tailored to manage input with one, two, three, and four channels, respectively. The outputs from the encoder are sample-specific features, denoted as $F \in \mathbb{R}^{C_{1}\times\frac{D}{16}\times\frac{H}{32}\times\frac{W}{32}}$, where $C_{1}$ represents the number of channels, and $D$, $H$, and $W$ indicate the depth, height, and width of the input, respectively.
In the decoder, each stage begins with an upsampling operation using a transposed convolution layer to gradually recover resolution while reducing the number of channels. The upsampled features are then concatenated with the corresponding outputs from the encoder and processed through two convolutional blocks. After the decoder stages, the output feature maps are passed through a segmentation head to produce segmentation maps, guided by a deep supervision strategy. The supervision signals are derived from a combination of Dice loss and cross-entropy loss to refine the training process.
%In the decoder, each stage begins with an upsampling operation using a transposed convolution layer to gradually recover resolution and reduce the number of channels. The upsampled features are then concatenated with the corresponding encoder stage outputs and processed through two convolutional blocks. Following the decoder stages, the output feature maps pass through a segmentation head to produce segmentation maps, supervised by a deep supervision strategy. The supervision signals are from a combination of Dice loss and cross-entropy loss to refine the training process.
The channel number for the multi-scale segmentation maps is set to the maximum number of classes across all tasks. For instance, in a scenario with datasets $S^{1}_{1}, S^{1}_{2}, S^{2}_{3}$ having class numbers of 5, 6, and 7 (including background classes), respectively, the output channel number is set to 7. Thanks to the prompt-based design (see Sections \ref{method: task prompt}), our method provides a significant advantage over multi-class models, which typically require up to 15 channels (\textit{i.e.}, 4+5+6), as these models often utilize binary cross-entropy loss, excluding the background class from the count.
%The channel number for the multi-scale segmentation maps is set to the maximum number of classes across all tasks. For example, considering a scenario with datasets ${S^{1}_{1}, S^{1}_{2}, S^{2}_{3}}$ having class numbers of 5, 6, and 7 (including background classes), respectively, the output channel number is set to 7. Thanks to the prompt-based design (introduced in Sections \ref{method: task prompt} and \ref{method: modal prompt}), our method offers a significant advantage over multi-class models, which typically require up to 15=4+5+6 channels (summing individual class counts) as these often use binary cross-entropy loss, excluding the background class from the count.

\subsection{Universal Task Prompt for Dynamic Task Priors} \label{method: task prompt}
We posit that there exist correlations among different segmentation tasks\cite{godau2021task}. 
Recognizing the complexity of manually crafting these correlations, we introduce a learnable prompt, termed the universal task prompt, to effectively describe them, promoting interaction and fusion among various task priors. The universal task prompt is defined as $F_{uni} \in \mathbb{R}^{K\times\frac{D_{3d}}{16}\times\frac{H_{3d}}{32}\times\frac{W_{3d}}{32}}$, where $K$ is a hyperparameter, and $D_{3d}$, $H_{3d}$, and $W_{3d}$ represent the depth, height, and width of 3D input data, respectively. 
A crucial aspect of training a universal network is ensuring the model is `aware' of the ongoing task during the feed-forward process. As a prompt-based model, MedUniSeg generates task-specific priors in a new manner (see Fig. \ref{fig: overview}).

Initially, it generates $N$ features by passing the concatenation of $F_{uni}$ and $F$ (the sample-specific features) through three convolutional blocks and splitting it along the channel dimension. This can be formally expressed as
\begin{equation}\label{eq1}
\{F_{task1},F_{task2},...,F_{taskN}\}=split(f(cat(F_{uni},F)))^{N},
\end{equation}
where $F_{taski}$ denotes the prompt features corresponding to the $i$-th task, $cat(\cdot,\cdot)$ represents the concatenation operation, $f(\cdot)$ denotes the feed-forward process, and $split(\cdot)^{N}$ divides the features along the channel dimension to yield $N$ features of identical shape.

Subsequently, we select the task-specific prior $F_{tp}$ from $\{F_{task1},F_{task2},...,F_{taskN}\}$ based on the Task ID of the current task. This selected feature $F_{tp}$ is concatenated with $F$ to form the input for the decoder. Notably, for 2D data, we perform interpolation on the sample-specific features and task-specific prior to ensure alignment with the shapes of the universal task prompt and sample-specific features, respectively. This method introduces task-related priors into the model at the end of the encoding process, enhancing the task-specific training of the entire decoder rather than limiting it to the final convolutional layers or the entire feed-forward process.

\subsection{Modal-specific Prompts for Modal Priors} \label{method: modal prompt}
As the number of modalities increases, optimization challenges arising from significant gaps among these modalities can hinder effective learning \cite{ye2024continual,huang2022modality,wang2020makes}. To address this issue, we introduce a strategy that enhances the model's ability to `aware' these modal gaps by incorporating modal-specific priors. This is achieved through a set of learnable modal prompts, denoted by $F_{mod}=\{ F_{mod_1}, F_{mod_2},..., F_{mod_M} \}$, where $F_{mod_M} \in \mathbb{R}^{l}$ represents the prompt corresponding to the modality with ID $M$, and $l$ is the length of each prompt.
The process begins with selecting the modal-specific prompt based on the modality ID of the input image. The selected prompt is then processed through the MMap module, which adapts the prompt to the input data's shape. The MMap module consists of a linear layer that maps the prompt from length $l$ to $144$, a reshaping operation that modifies this mapped prompt from $144$ to $12 \times 12$ for 2D images or $4 \times 6 \times 6$ for 3D images, and a linear interpolation that resamples the resized prompt to match the shape of the input image.
Unlike Hermes \cite{gao2024training}, which integrates the modal prior at multiple stages across the encoding and decoding processes, we introduce the prior only once at the beginning of the encoder, carefully controlling the number of parameters involved. This results in an approximate increase of 80K parameters to accommodate the prompts for nine modalities. The design principle behind the prompt's introduction is to address differences as they arise, which is particularly crucial for modalities at the start of the encoding process.
%The design principle behind the prompt's introduction position is to reply to differences as they initially appear. For modalities, these differences are at the very start of the encoding process.
Ultimately, the input data for the encoder are formulated by combining the input images with the modal priors, as shown below:
\begin{equation}\label{eq2}
Input = I + MMap(select(F_{mod}, m),
\end{equation}
where $m$ is the modal ID of the input image $I$, $select(F_{mod}, m)$ selects the corresponding $m$-th modal-specific prompt from the set $F_{mod}$, and $MMap(\cdot)$ processes this prompt through the MMap module. 

The reason for employing individual prompts for each modality, rather than a universal prompt as in our task prior strategy, stems from the fact that the upstream dataset is multi-modal but unpaired, leading to negligible correlations between different modal data.

\subsection{Transfer Learning}
After training MedUniSeg on the upstream dataset, we transfer the pre-trained encoder-decoder along with the randomly initialized segmentation head to the downstream task. Additionally, the branch responsible for generating the modal prior is also transferred. We freeze the corresponding modal-specific prompt to preserve its learned characteristics, while the linear layer of the MMap module remains learnable to focus on mapping the specific modal prompt. The model is fine-tuned in a fully supervised manner to minimize the sum of the Dice loss and cross-entropy loss.

%=====================================================================================================================================
%===============================================================Datasets==============================================================
%=====================================================================================================================================
\begin{table*}[t]
  \centering
  \caption{Details of 17 upstream datasets and six downstream datasets.}
  \setlength{\tabcolsep}{2pt}
  \resizebox{2.0\columnwidth}{!}{
    \begin{tabular}{c|cccccccc|cc|c|c|c|c|c|c|c|cc|c|c|c|c}
    \hline \hline
    \multirow{3}[6]{*}{Dataset} & \multicolumn{17}{c|}{Upstream}                                                                                                        & \multicolumn{6}{c}{Downstream} \\
\cmidrule{2-24}          & \multicolumn{8}{c|}{CT}                                       & \multicolumn{2}{c|}{MRI} & CT\&PET & Dermoscopic & Fundus & Path. & Ultrasound & \multicolumn{1}{c|}{X-ray} & Endoscope & \multicolumn{2}{c|}{CT} & MRI   & X-ray & Fundus & Path. \\
\cmidrule{2-24}          & Liver & Kidney & HepaV & Pancreas & Colon & Lung  & Spleen & VerSe20 & Prostate & BraTS21 & AutoPET22 & ISIC18 & REFUGE2 & GlaS  & BUSI  & QaTav2 & Polyp & BTCV  & COVID-19-20 & VS    & SIIM  & IDRID & SegPC \\
    \midrule
    Target & Organ\&Tumor & Organ\&Tumor & Organ\&Tumor & Organ\&Tumor & Tumor & Tumor & Organ & Vertebrae & Organ & Tumor & Tumor & Lesion & Tissue & Tissue & Tumor & Lesion & Lesion & Organ & Lesion & Tumor & Lesion & Lesion & Cell \\
    Train & 104   & 168   & 242   & 224   & 100   & 50    & 32    & 171   & 91    & 1000  & 400   & 2694  & 1600  & 85    & 623   & 7145  & 1450  & 21    & 159   & 193   & 5048  & 54    & 298 \\
    Test  & 27    & 42    & 61    & 57    & 26    & 13    & 9     & 43    & 25    & 251   & 101   & 1000  & 400   & 80    & 157   & 2113  & 798   & 9     & 40    & 49    & 1372  & 27    & 199 \\
 \hline \hline
    \end{tabular}%
    }
  \label{tab: dataset}%
\end{table*}%

\begin{table}[t]
  \centering
  \caption{Patch sizes and batch sizes for all fine-tuning models on the six downstream datasets.}
  \setlength{\tabcolsep}{2pt}
  \resizebox{1.0\columnwidth}{!}{
    \begin{tabular}{c|cccccc}
    \hline \hline
    Dataset	    &BTCV	&COVID-19-20 &VS	&SIIM	&IDRID	&SegPC \\
    \hline
    Batch Size	&2	    &2	          &2	    &12	    &12	    &12 \\  
    Patch Size	&1$\times$48$\times$192$^{2}$	&1$\times$64$\times$192$^{2}$	&1$\times$48$\times$192$^{2}$	&3$\times$1$\times$512$^{2}$	&3$\times$1$\times$512$^{2}$   &3$\times$1$\times$512$^{2}$ \\
 \hline \hline
    \end{tabular}%
    }
  \label{tab: setting}%
\end{table}%

\section{Datasets}
We categorize the datasets used in this study into two groups: an upstream dataset and six downstream datasets.

\textbf{Upstream Dataset.}
To train our MedUniSeg model and compare it against other universal and single-task models, we collected an upstream dataset comprising 17 public sub-datasets, each annotated with specific targets. 
The \textbf{Liver} dataset, derived from LiTS \cite{bilic2023liver}, contains contrast-enhanced abdominal CT scans annotated with livers and liver tumors. 
The \textbf{Kidney} dataset, sourced from KiTS \cite{heller2021state}, includes CT scans of kidney cancer patients who underwent nephrectomy, annotated with kidneys and kidney tumors. 
The \textbf{HepaV}, \textbf{Pancreas}, \textbf{Colon}, \textbf{Lung}, and \textbf{Spleen} datasets were taken from the Medical Segmentation Decathlon (MSD) Challenge \cite{antonelli2022medical}, covering segmentation tasks for hepatic vessels, hepatic tumors, pancreases, pancreas tumors, colon tumors, lung tumors, and spleens, respectively. 
The \textbf{VerSe20} dataset \cite{sekuboyina2021verse} provides segmentation annotations of vertebrae, and we utilized its binary form, merging all foreground classes into a single category.
The \textbf{Prostate} dataset combines the NCI-ISBI 2013 dataset \cite{nicholas2015nci}, I2CVB dataset \cite{lemaitre2015computer}, and PROMISE12 dataset \cite{litjens2014evaluation} for multi-domain prostate segmentation. 
The \textbf{BraTS21} dataset \cite{baid2021rsna} annotates brain tumors across four MRI modalities (T1, T1-weighted, T2-weighted, and T2-FLAIR), providing segmentation for peritumoral edematous/invaded tissue, the necrotic tumor core, and the Gd-enhancing tumor. 
The \textbf{AutoPET22} dataset \cite{gatidis2022whole} offers PET scans with whole-body tumor annotations. 
The \textbf{ISIC18} dataset \cite{codella2019skin} contains skin lesion annotations, classifying images as cancerous or non-cancerous. 
The \textbf{REFUGE2} dataset \cite{fang2022refuge2} provides annotations for glaucoma classification, optic disc/cup segmentation, and fovea localization; we used only the segmentation annotations. 
The \textbf{GlaS} dataset \cite{sirinukunwattana2017gland} labels H\&E-stained colon tissue images as malignant or benign. 
The \textbf{BUSI} dataset \cite{al2020dataset} includes images categorized as normal, benign, or malignant, with tumor annotations for the latter two categories. 
The \textbf{QaTav2} dataset \cite{degerli2022osegnet} focuses on segmenting COVID-19 infected regions. 
The \textbf{Polyp} dataset \cite{fan2020pranet} consists of five sub-datasets, including Kvasir \cite{jha2020kvasir}, CVC-ClinicDB \cite{bernal2015wm}, CVC-ColonDB \cite{tajbakhsh2015automated}, ETIS \cite{silva2014toward}, and CVC-300 \cite{bernal2012towards}, for polyp segmentation.

\textbf{Downstream Datasets.}
To evaluate the transfer capabilities of well-trained universal models, supervised models, and self-supervised models, we employed six 2D or 3D segmentation datasets. 
The \textbf{BTCV} dataset \cite{landman2015miccai} provides annotations for 13 abdominal organs, including the spleen, right and left kidneys, gallbladder, esophagus, liver, stomach, aorta, inferior vena cava, portal vein, splenic vein, pancreas, and adrenal glands. 
The \textbf{COVID-19-20} dataset includes annotations of COVID-19 lung CT lesions \cite{roth2022rapid}. 
The \textbf{VS} dataset \cite{shapey2021segmentation} contains annotations for vestibular schwannomas. 
The \textbf{SIIM} dataset \cite{siim-acr-pneumothorax-segmentation} provides segmentation annotations for pneumothorax. To address the imbalance of normal and lesion training samples, we followed \cite{huang2021gloria} and balanced the dataset by reducing the number of normal training samples until it was the same as the number of lesion training samples. 
The \textbf{IDRID} dataset \cite{porwal2018indian} was used to offer annotations for hemorrhages and hard exudates. 
The \textbf{SegPC} dataset \cite{bozorgpour2021multi} includes annotations for cytoplasm and nucleus segmentation in myeloma plasma cells.

Detailed information about each dataset is provided in Table \ref{tab: dataset}. For data splits, we adhered to established protocols whenever available, following the official data splits for datasets like ISIC18 or widely accepted splits such as those for the BTCV dataset. For datasets lacking pre-defined splits, we randomly divided the available data using an 80:20 ratio for training and testing, respectively.

%=====================================================================================================================================
%============================================================Experiments==============================================================
%=====================================================================================================================================

% \begin{figure}[t]
%   \centering
%   \includegraphics[width=1.0\linewidth]{Image/modal_prompt.pdf}
%    \caption{T-SNE visualization of the nine modality features obtained from MedUniSeg w/o modal priors or MedUniSeg. The black arrows are used to highlight the significant differences.
%    }
% \label{fig: visual modal prompt}
% \end{figure}

\section{Experiments}
\subsection{Implementations}
We implemented both joint training on the upstream dataset and fine-tuning on six downstream datasets using the nnUNet framework.

\textbf{Universal training.}
The Stochastic Gradient Descent (SGD) optimizer was utilized, starting with an initial learning rate of 0.01. Batch sizes varied according to data dimensions: 12 for 2D data and 2 for 3D data. The patch sizes were set to $3 \times 1 \times 512 \times 512$ for 2D data and $1 \times 64 \times 192 \times 192$ for 3D data. The training was designed to run for a maximum of 1,000 epochs, with each dataset allocated 50 iterations per epoch, totaling 850 iterations.
%The training process was designed to run for a maximum of 1,000 epochs. During each epoch, every dataset was allocated 50 iterations, summing up to a total of 850 iterations per epoch.

\textbf{Fine-tuning.}
%Swin UNETR-based 64,192,192
For fine-tuning, we continued using the nnUNet framework. The batch size and patch size for each downstream dataset were detailed in Table \ref{tab: setting}. The initial learning rate was remained at 0.01, with a maximum of 25,000 training iterations for most datasets. For the SIIM dataset, we extended this to 100,000 iterations to ensure convergence. Each method was executed three times for each dataset, and average results were reported.

%We continued using the nnUNet framework, which adaptively selected batch sizes and patch sizes appropriate for each dataset. The initial learning rate was maintained at 0.01, with 25,000 maximum training iterations for most datasets. An exception was made for the SIIM dataset, where 100,000 iterations were set to ensure the model reached convergence. For each dataset, we ran each method three times and reported the average results.

Detailed pre-processing steps for each dataset were provided in our publicly accessible code.

\subsection{Evaluation Metrics}
The Dice similarity coefficient (Dice, \%) was used as the primary metric for evaluating model performance. For datasets with multiple foreground categories, we computed the mean Dice score over these categories to reflect overall performance. In contrast, for the SIIM dataset, which exhibits significant class imbalance (290 normal vs. 1,082 lesion images), we utilized the weighted Dice (WDice) to ensure a fair evaluation. WDice is calculated as follows: 
\begin{equation}\label{eq3}
WDice= w_{0} \times D_{0} + w_{1} \times D_{1},
\end{equation}
where $w_{0}$ and $w_{1}$ are the weights assigned to the normal and lesion categories, respectively. Both weights are inversely proportional to the frequency of each class, ensuring equitable contribution from both categories to the evaluation metric. Here, $D_{0}$ and $D_{1}$ denote the mean Dice scores for the normal and lesion images, respectively.

\begin{table*}[t]
  \centering
  \caption{Performance of single-task models and universal models on 17 datasets. Dice scores (\%) are reported for each dataset, with 3D mean Dice (\%) calculated for all 3D datasets, 2D mean Dice (\%) for all 2D datasets, and mean Dice (\%) for all datasets. The best results for each dataset are highlighted in bold.}
    \setlength{\tabcolsep}{2pt}
    \resizebox{2.0\columnwidth}{!}{
    \begin{tabular}{c|ccccccccccc|cccccc|ccc}
    \hline\hline
    Method & Liver & Kidney & HepaV & Pancreas & Colon & Lung  & Spleen & VerSe20 & Prostate & BraTS21 & AutoPET22 & ISIC18 & REFUGE2 & GlaS  & BUSI  & QaTav2 & Polyp & 3D Mean & 2D Mean & Mean \\
    \hline
    \multicolumn{21}{c}{Single-task Model} \\
    nnFormer \cite{zhou2023nnformer} & 70.7  & 80.0  & 61.3  & 57.9  & 18.8  & 66.8  & 92.2  & 84.3  & 87.0  & 82.0  & 61.0  & 87.8  & 90.2  & 90.5  & 74.7  & 77.8  & 60.5  & 69.3  & 80.3  & 73.2 \\
    MiT \cite{xie2022unimiss}   & 71.5  & 76.8  & 63.8  & 58.1  & 32.0  & 60.7  & 95.7  & 85.3  & 85.9  & 82.7  & 59.7  & 88.8  & 90.6  & 90.5  & 77.0  & 79.0  & 70.5  & 70.2  & 82.7  & 74.6 \\
    CoTr \cite{xie2021cotr}  & 74.7  & 85.1  & 67.2  & 65.8  & 33.8  & 66.9  & 95.2  & 87.1  & 88.0  & 82.9  & 58.8  &88.0	&89.1	&89.9	&77.7	&79.6	&77.0	&73.2	&83.5	&76.9 \\
    UXNet \cite{lee2023d} & 75.4  & 82.2  & 67.3  & 59.4  & 39.8  & 59.5  & 95.7  & 87.1  & 88.8  & 84.3  & 68.2  & 88.8  & 90.5  & 88.9  & 78.6  & 79.2  & 73.3  & 73.4  & 83.2  & 76.9 \\
   Swin UNETR \cite{tang2022self} &74.8	&82.6	&68.3	&63.8	&41.1	&71.5	&96.2	&86.6	&88.7	&84.2	&59.2	&88.5	&91.2	&90.1	&76.4	&78.3	&71.7	&74.3	&82.7	&77.3 \\
    UCI \cite{guan2023unpaired}   & 78.2  & 85.0  & 67.6  & 63.7  & 40.4  & 68.1  & 95.9  & 86.6  & 88.7  & 84.1  & 64.2  & 89.1  & 90.5  & 90.0  & 75.3  & 78.5  & 71.7  & 74.8  & 82.5  & 77.5 \\
    UKAN \cite{li2024u} & 76.0  & 86.6  & 70.0  & 65.2  & 47.0  & 66.1  & 96.0  & 86.5  & 89.3  & 83.8  & 66.5  & 88.4  & \textbf{91.5} & 91.2  & 79.1  & 79.6  & 73.6  & 75.7  & 83.9  & 78.6 \\
    U-Mamba \cite{ma2024u} & 77.5  & 86.2  & 70.4  & 70.5  & 47.0  & 68.2  & 95.8  & 87.2  & 88.6  & \textbf{84.6}  & 64.8  & 88.7  & 91.1  & 90.8  & 78.1  & \textbf{80.6}  & \textbf{77.8}  & 76.4  & 84.5  & 79.3 \\
    nnUNet \cite{isensee2021nnu} & 77.2  & \textbf{87.5}  & 69.6  & 68.8  & 49.0  & 68.4  & 96.2  & \textbf{87.2}  & 89.4  & 84.4  & 64.6  & 88.4  & 90.8  & 90.4  & 79.1  & 80.1  & 77.6  & 76.6  & 84.4  & 79.3 \\
    \hline
    \multicolumn{21}{c}{Universal Model} \\
    Universal Model \cite{liu2023clip} &75.2	&85.8	&69.5	&63.9	&49.9	&61.1	&96.3	&86.0	&89.3	&83.6	&67.4	&88.8	&90.7	&90.6	&79.7	&79.6	&74.3	&75.3	&83.9	&78.3 \\
    Hermes \cite{gao2024training} &75.6	&84.1	&69.1	&66.8	&48.8	&68.8	&96.4	&86.1	&88.6	&83.8	&67.5	&89.3	&90.3	&90.2	&77.5	&79.2	&73.6	&76.0	&83.4	&78.6 \\
    DoDNet \cite{zhang2021dodnet} & 76.9  & 87.3  & 69.9  & 69.8  & 53.0  & 65.8  & 96.4  & 86.1  & 89.2  & 83.0  & 62.1  & 89.3  & 90.9  & 90.9  & 79.4  & 78.3  & 75.4  & 76.3  & 84.0  & 79.0 \\
    CCQ \cite{liu2023ccq}  & 76.6  & 86.6  & \textbf{70.5}  & 68.9  & \textbf{54.8}  & 69.7  & 96.3  & 86.2  & 89.4  & 83.3  & 61.8  & 88.9  & 90.7  & 90.7  & 79.1  & 78.4  & 76.3  & 76.7  & 84.0  & 79.3 \\
    UniSeg \cite{ye2023uniseg} & 79.0  & 87.0  & 70.4  & 69.8  & 53.5  & 69.0  & 96.4  & 86.1  & 89.9  & 83.6  & 67.7  & \textbf{89.4}  & 91.3  & 90.9  & 79.8  & 78.6  & 76.2  & 77.5  & 84.3  & 79.9 \\
    MedUniSeg & \textbf{79.9}  & 86.9  & 70.2  & \textbf{71.0}  & 54.2  & \textbf{72.6}  & \textbf{96.4}  & 86.3  & \textbf{89.9} & 83.5  & \textbf{68.7}  & 89.2  & 91.3  & \textbf{91.6}  & \textbf{80.4}  & 78.8  & 77.5  & \textbf{78.1}  & \textbf{84.8}  & \textbf{80.5} \\
    \hline
    MedUniSeg* &79.9	&89.0	&70.2	&71.0	&54.2	&72.6	&96.4	&86.8	&89.9	&84.4	&68.7	&89.2	&91.3	&91.6	&80.4	&79.9	&78.1	&78.5	&85.1	&80.8 \\
    \hline\hline
    \end{tabular}%
    }
  \label{tab: seg comp}%
\end{table*}%

\begin{table*}[t]
  \centering
  \caption{Performance of ten self-supervised models, five supervised models, and two training from scratch (TFS) models on six downstream datasets, utilizing 20\%, 50\%, and 100\% of the training data. For 3D models, the 2D data are regarded as pseudo 3D data with a depth of one. A dash $-$ presents that the model could not be trained on the dataset. For universal models, a dagger $\dag$ means the use of official pre-trained weights. Dice scores (\%) are reported for each dataset. All results represent the average of three independent runs, with the best performance for each dataset highlighted in bold. }
      \setlength{\tabcolsep}{4pt}
    \resizebox{2.0\columnwidth}{!}{
    \begin{tabular}{c|c|ccc|ccc|ccc|ccc|ccc|ccc}
    \hline\hline
    \multirow{3}[6]{*}{Method} & \multirow{3}[6]{*}{Pre-training Data} & \multicolumn{9}{c|}{3D}                                               & \multicolumn{9}{c}{2D} \\
\cmidrule{3-20}          &       & \multicolumn{3}{c|}{BTCV (CT)} & \multicolumn{3}{c|}{COVID-19-20 (CT)} & \multicolumn{3}{c|}{VS (MRI)} & \multicolumn{3}{c|}{SIIM (X-ray)} & \multicolumn{3}{c|}{IDRID (Fundus)} & \multicolumn{3}{c}{SegPC (Path.)} \\
\cmidrule{3-20}          &       & 20\%  & 50\%  & 100\% & 20\%  & 50\%  & 100\% & 20\%  & 50\%  & 100\% & 20\%  & 50\%  & 100\% & 20\%  & 50\%  & 100\% & 20\%  & 50\%  & 100\% \\
    \hline
    MG \cite{zhou2021models} & 3D CT & 50.2  & 66.1  & 77.1  & 59.7  & 62.2  & 63.3  & 81.3  & 88.6  & 85.9  & 41.1 & 48.8  & 52.2  & 24.1  & 29.2  & 22.9  & 70.8  & 75.1  & 77.5 \\
    GVSL \cite{he2023geometric} & 3D CT & 31.4  & 69.8  & 79.5  & 54.5  & 55.4  & 56.5  & 86.9  & 87.9  & 91.0  &  43.1    & 49.1  & 52.9  & 39.4  & 47.3  &  49.3     & 73.1  & 77.2  & 80.3 \\
    DeSD \cite{ye2022desd} & 3D CT & 69.5  & 79.5  & 83.3  & 62.8  & 67.1  & 68.3  & 91.1  & 91.4  & 92.2  & 39.1  & 42.2  & 46.0  & 47.7  & 60.7  & 59.4  & 75.4  & 79.3  & 80.8 \\
    SMIT \cite{jiang2022self} & 3D CT & 56.7  & 72.5  & 80.6  & 57.3  & 58.7  & 62.1  & 90.3  & 91.5  & 92.2  & -     & -     & -     & -     & -     & -     & -     & -     & - \\
    Swin UNETR \cite{tang2022self} & 3D CT &58.8 &74.1	&80.7	&58.0	&60.6	&63.7 &89.6	&89.1	&90.0    & -     & -     & -     & -     & -     & -     & -     & -     & - \\
    VoCo \cite{wu2024voco} &3D CT &68.9	&78.7	&83.4	&62.0	&64.9	&67.6	&91.1	&91.9	&92.7	&-	&-	&-	&-	&-	&-	&-	&-	&- \\
    BT \cite{kang2023benchmarking} & 2D Path. & -     & -     & -     & -     & -     & -     & -     & -     & -     & 44.4  & 51.2  & 54.2  & 49.0  & 56.5  & 58.2  & 76.0  & 79.7  & 80.2 \\
    PCRLv2 (CheXpert) \cite{zhou2023unified} & 2D X-ray & -     & -     & -     & -     & -     & -     & -     & -     & -     & 38.7  & 47.6  & 49.7  & 35.2  & 39.1  & 51.3  & 76.0  & 78.8  & 79.4 \\
    MedKLIP \cite{wu2023medklip}	& 1D Report, 2D X-ray	&-	&-	&-	&-	&-	&-	&-	&-	&-	&48.9	&53.0	&54.2	&41.2	&47.5	&51.6	&73.0	&77.2	&78.2		 \\
    UniMiSS \cite{xie2022unimiss} & 2D X-ray, 3D CT & 66.4  & 76.7  & 81.2  & 60.7  & 64.1  & 65.8  & 89.9  & 90.8  & 91.4  & 46.0  & 52.7  & 54.8  & 51.4   &  61.5  & 63.5   & 73.8  & 78.8  & 80.7 \\
    \hline
    UniSeg$\dag$ \cite{ye2023uniseg} & 3D CT, 3D MRI, 3D PET & 71.4	&79.7	&84.6	&68.6	&70.9	&72.0	&91.1	&92.1	&92.9	&50.8	&56.2	&58.3	&53.9	&62.5	&63.9	&75.3	&80.8	&82.5 \\
    Universal Model$\dag$ \cite{liu2023clip}  & 3D CT &61.9	&76.1	&79.9	&61.0	&62.8	&66.1	&91.2	&91.3	&92.3	&-	&-	&-	&-	&-	&-	&-	&-	&- \\
    \hline
    2D Backbone  & N/A     & -     & -     & -     & -     & -     & -     & -     & -     & -     & 43.9  & 53.2  & 55.6  & 53.4  & 61.5  & 62.8  & 74.8  & 79.1  & 82.0 \\
    3D Backbone  & N/A     & 66.2  & 77.9  & 83.1  & 61.2  & 61.6  & 65.0  & 89.7  & 89.9  & 90.7  & 45.5  & 54.2  & 55.7  & 52.9  & 61.3  & 62.8  & 75.0  & 79.4  & 82.1 \\
    Universal Model \cite{liu2023clip} & Nine Modalities &71.0	&79.5	&84.2	&65.6	&66.0	&69.6	&90.6	&91.1	&91.8	&50.3	&56.7	&59.1	&54.8	&63.1	&64.3	&77.5	&82.1	&83.4 \\
    Hermes \cite{gao2024training}	&Nine Modalities	&68.0	&77.6	&83.8	&63.9	&65.8	&67.2	&90.2	&91.4	&91.8	&50.0	&56.6	&58.6	&54.6	&63.0	&64.5	&76.3	&81.3	&82.9 \\
    DoDNet \cite{zhang2021dodnet} & Nine Modalities & 70.9  & 78.9  & 83.8  & 67.9  & 71.3  & 71.9  & 91.7  & 92.1  & 93.0  & 48.8  & 56.3  & 58.8  & 54.6  & 62.8  & 64.2  & 77.3  & 81.8  & 83.0 \\
    CCQ \cite{liu2023ccq} & Nine Modalities & 70.9  & 79.4  & 84.1  & 67.5  & 69.1  & 71.9  & 91.6  & 92.0  & 92.2  & 49.8  & 56.6  & 58.9  & 54.0  & 62.8  & 64.1  & 77.6  & 81.9  & 83.1 \\
    UniSeg \cite{ye2023uniseg} & Nine Modalities & 71.4  & 79.5  & 84.4  & \textbf{69.1}  & 71.5  & 72.3  & 91.7  & 92.6  & 92.8  & 51.5  & 56.7  & 58.7  & 55.4  & 63.2  & 64.5  & 78.2  & 82.3  & 83.3 \\
    MedUniSeg & Nine Modalities & \textbf{71.8}  & \textbf{80.2}  & \textbf{84.6}  & 68.8  & \textbf{71.8} & \textbf{72.5}  & \textbf{92.3}  & \textbf{93.2}  & \textbf{94.0}  & \textbf{52.2}  & \textbf{57.0}  & \textbf{59.8}  & \textbf{55.5}  & \textbf{63.4}  & \textbf{64.9}  & \textbf{78.6}  & \textbf{82.7}  & \textbf{83.7} \\
  \hline\hline
    \end{tabular}%
    }
  \label{tab: fine-tuning}%
\end{table*}%

\subsection{Comparing to Single-task and Universal Models} \label{exp: upstream}
We compared our MedUniSeg with nine single-task models and five universal models. The single-task models include nnFormer \cite{zhou2023nnformer}, MiT \cite{xie2022unimiss},  CoTr \cite{xie2021cotr},Swin UNETR \cite{tang2022self}, UXNet \cite{lee2023d}, UCI \cite{guan2023unpaired}, UKAN \cite{li2024u}, U-Mamba (U-Mamba\_Bot) \cite{ma2024u}, and nnUNet \cite{isensee2021nnu}. The universal models consist of Universal Model \cite{liu2023clip}, Hermes \cite{gao2024training}, DoDNet \cite{zhang2021dodnet}, CCQ \cite{liu2023ccq}, and UniSeg \cite{ye2023uniseg}. For single-task models, each dataset was used for individual model training, employing both 3D and 2D versions to address corresponding tasks. To ensure a fair comparison, all single-task models were trained for a maximum of 1,000 epochs, each containing 50 iterations. The patch size for these models was $64\times192\times192$ for 3D data and $512\times512$ for 2D data. The backbones of the competing universal models and our MedUniSeg remained consistent across comparisons. All models were trained from scratch.

\begin{figure*}[t]
  \centering
  \includegraphics[width=1.0\linewidth]{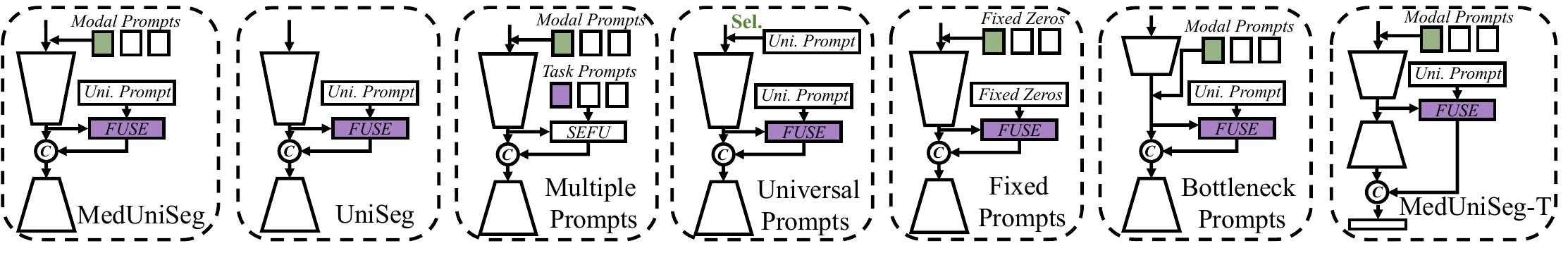}
   \caption{Schematic representation of MedUniSeg, UniSeg, Multiple Prompts, Universal Prompts, Fixed Prompts, Bottleneck Prompts, and MedUniSeg-T. Multiple Prompts utilizes multiple task-specific and modal-specific prompts. Universal Prompts adopts a universal modal prompt and a universal task prompt. Fixed Prompts initializes with zero prompts, remaining unchanged. Bottleneck Prompts incorporates both priors at the bottleneck of the encoder. MedUniSeg-T introduces the task-related prompt at the end of the decoder. The selection and fusion (SEFU) module first selects a modal-specific prompt and then fuses the features with the prompt. The $Sel.$ operation is used to extract the modal-specific prior from the universal prompt generated by the MMap module. Task-related information is highlighted in purple, while modal-related information is highlighted in green.
   }
\label{fig: variants}
\end{figure*}

\begin{table*}[t]
  \centering
  \caption{Performance of baseline, six variants, and our MedUniSeg. The baseline refers to our encoder-decoder backbone trained independently on each dataset. We compare the 3D mean Dice (\%), 2D mean Dice (\%), and mean Dice (\%) across all models.}
    \setlength{\tabcolsep}{4pt}
    \resizebox{2.0\columnwidth}{!}{
    \begin{tabular}{ccccccccc}
    \hline\hline
    Method & Baseline &UniSeg & Multiple Prompts &Universal Prompts &Fixed Prompts &Bottleneck Prompts &MedUniSeg-T &MedUniSeg \\ 
    \hline
    3D Mean & 76.6 &77.5 &77.1 &76.6 &77.5 &77.7 &77.2 &78.1 \\
    2D Mean &84.4 &84.3 &84.8 &84.9 &84.4 &84.4 &84.4 &84.8 \\
    Mean    &79.3 &79.9 &79.8 &79.5 &79.9 &80.1 &79.7 &80.5 \\
    \hline\hline
    \end{tabular}%
    }
  \label{tab: ablation}%
\end{table*}%

The results presented in Table \ref{tab: seg comp} lead to two main conclusions: 
First, Transformer-based methods, such as nnFormer, MiT, Swin UNETR, UXNet, and UCI, generally underperform compared to CNN-based methods like nnUNet in segmentation tasks, particularly for 3D data. This observation prompted us to favor a pure CNN-based model for both 2D and 3D universal segmentation over Transformer-based models.
Additionally, nnUNet and U-Mamba demonstrated superior generalization performance compared to other single-task models, with a 0.7\% improvement in average performance over the third-best model, UKAN. Considering both performance and model size (U-Mamba: $\sim$48.2M vs. nnUNet: $\sim$31.2M), nnUNet was selected as the backbone for the universal models. 
Second, the increasing challenge of addressing segmentation tasks over various modalities, regions, and domains revealed that recent advanced universal models often struggle to achieve satisfactory performance, typically scoring lower average Dice scores than the baseline, \textit{i.e.}, nnUNet. 
In contrast, our UniSeg and MedUniSeg models demonstrate improved performance, achieving mean Dice score increases of 0.6\% and 1.2\% over the baseline, respectively. Furthermore, MedUniSeg attains a 1.5\% improvement for 3D tasks and 0.4\% for 2D tasks.
In summary, our MedUniSeg achieves the best generalization performance across 17 segmentation tasks, effectively addressing multiple tasks with a single model while consistently outperforming its baseline on most tasks.

\begin{figure*}[t]
  \centering
  \includegraphics[width=1.0\linewidth]{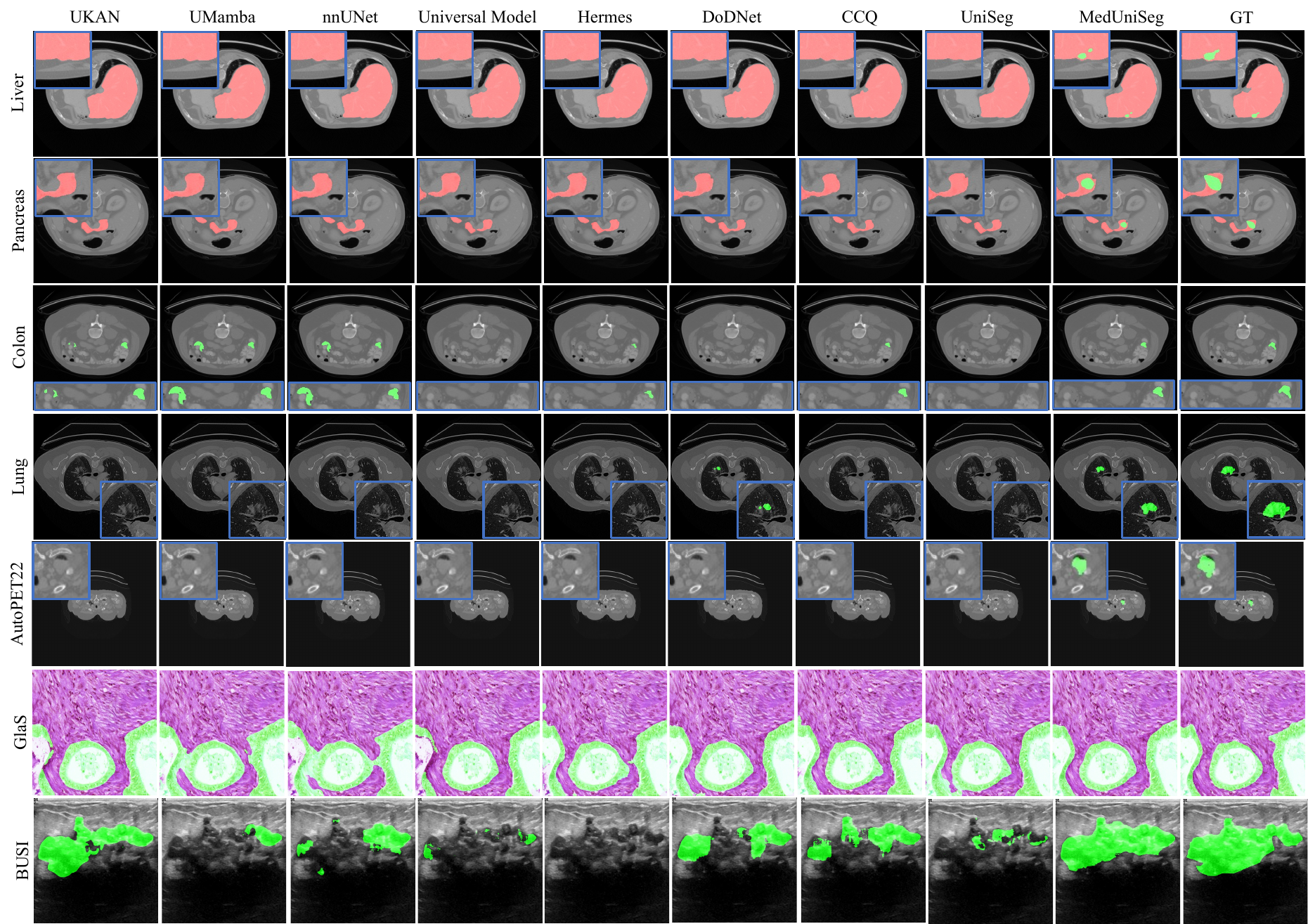}
   \caption{Visualization of segmentation results obtained from UKAN, UMamba, nnUNet, Universal Model, Hermes, DoDNet, CCQ, UniSeg, and MedUniSeg, along with the ground truths (GTs) on seven datasets. Organs are depicted in red, while tumors and lesions are shown in green. Blue rectangles highlight the differences among the models.
   }
\label{fig: upstream visualization}
\end{figure*}

\begin{figure*}[t]
  \centering
  \includegraphics[width=1.0\linewidth]{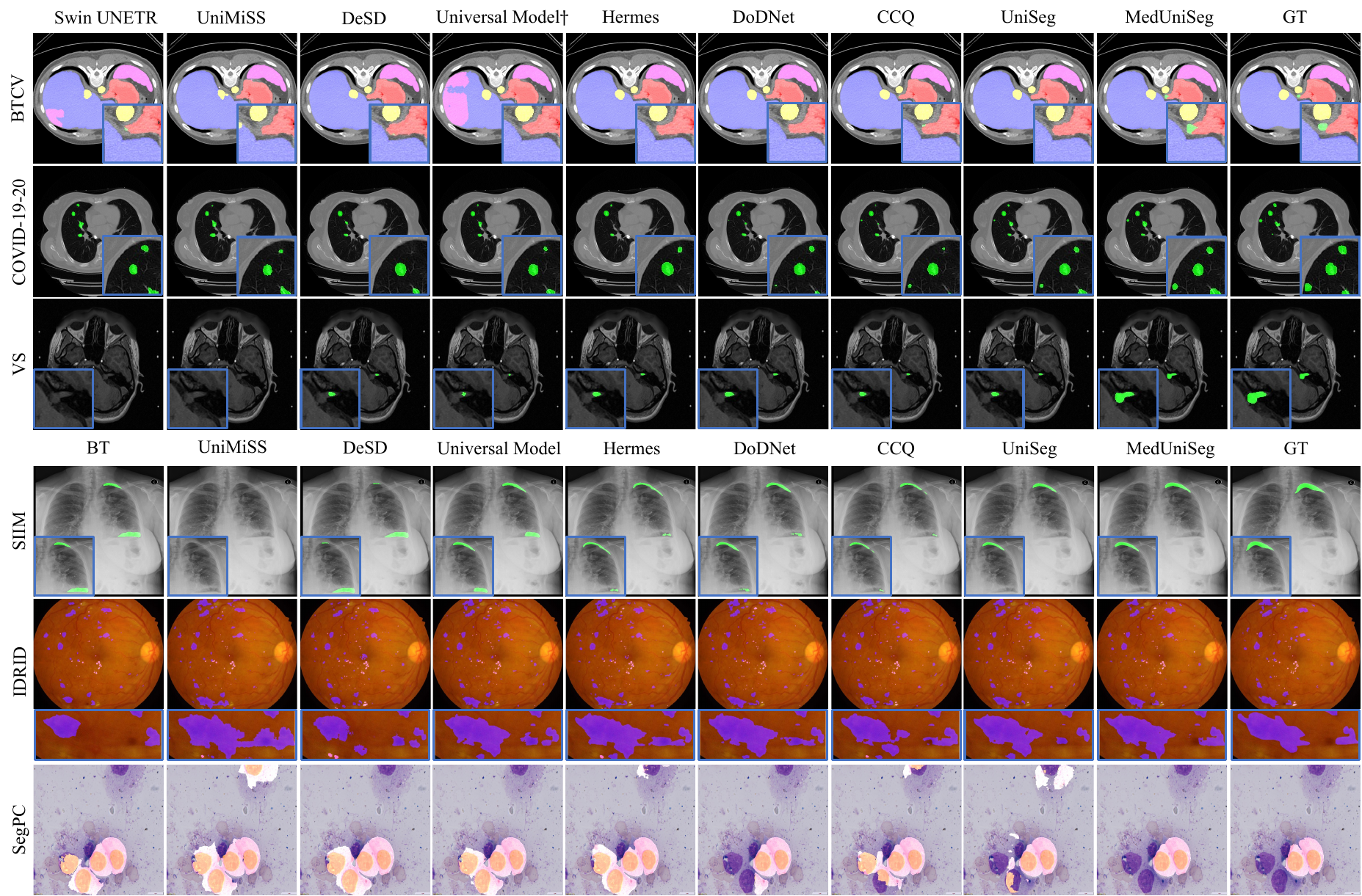}
   \caption{Visualization of segmentation results obtained from Swin UNETR, BT, UniMiSS, DeSD, Universal Model, Universal Model$\dag$, Hermes, DoDNet, CCQ, UniSeg, and MedUniSeg, along with the ground truths (GTs) on six datasets. Blue rectangles highlight the differences among the models.
   }
\label{fig: downstream visualization}
\end{figure*}

\begin{figure}[t]
  \centering
  \includegraphics[width=1.0\linewidth]{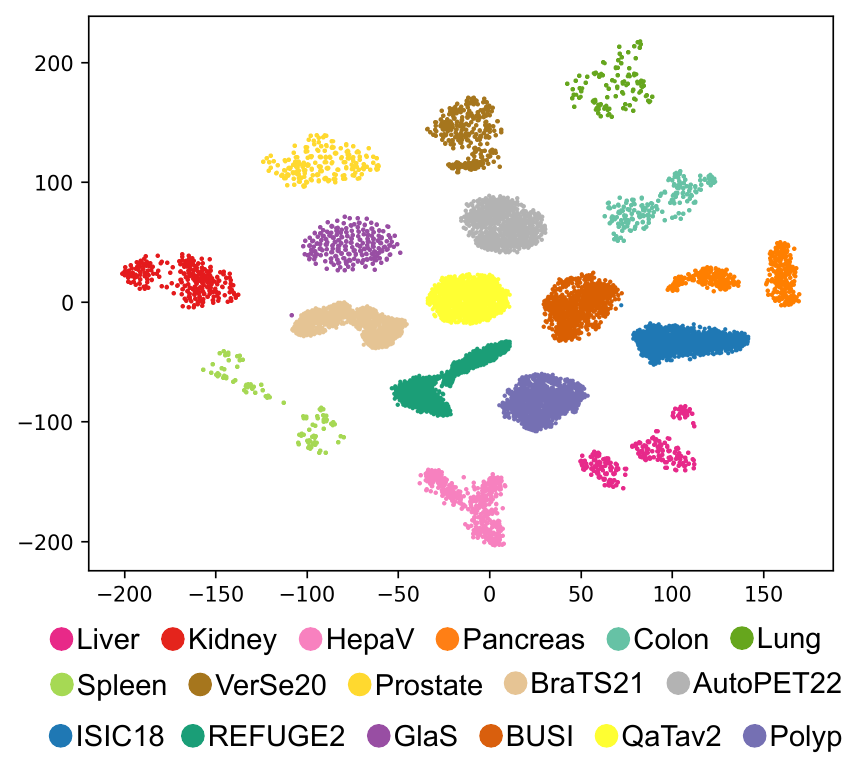}
   \caption{T-SNE of 17 task-specific priors, illustrating the distributions among the tasks. 
   }
\label{fig: visual task prompt}
\end{figure}

\subsection{Performance Improvement using LoRA}
Table \ref{tab: seg comp} shows that MedUniSeg generally outperforms the baseline nnUNet but slightly underperforms on five tasks: Kidney, VerSe20, BraTS21, QaTav2, and Polyp. 
To address these performance gaps, we enhanced MedUniSeg by freezing the trained models and integrating learnable LoRA modules into convolutional layers. The rank and alpha were set to 32 and 64, respectively.
Moreover, we introduced new deconvolutional layers and segmentation heads to produce residual outputs. This modified model, referred to as MedUniSeg*, was retrained on the five under-performing tasks, updating only the newly added modules. All other configurations, such as 1000 epochs and 50 iterations per epoch, remain consistent with the upstream training. As shown in Table \ref{tab: seg comp}, MedUniSeg* demonstrates performance improvements over MedUniSeg on all five tasks, with Dice score gains of 2.1\%, 0.5\%, 0.9\%, 1.1\%, and 0.6\%, respectively, while increasing the parameter count by approximately 6.8M. Furthermore, when comparing MedUniSeg* to nnUNet, MedUniSeg* outperforms nnUNet on 14 tasks, matches its performance on one task, and shows slightly lower performance on only two tasks.

\begin{table*}[t]
  \centering
  \caption{Results of the FUSE module with varying block numbers and channel numbers. Here, $\#Bi$ means the FUSE module with $i$ blocks, while $Ci$ indicates that the middle layer of the FUSE module reduces the channel count to $1/i$ of the original. The best results are highlighted in bold.}
    \resizebox{2.0\columnwidth}{!}{
    \begin{tabular}{cccccc|ccccc}
    \hline\hline
    Method & \#B1  & \#B2, C4 & \#B3, C4 & \#B4, C4 & \#B5, C4 & \#B3, C1 & \#B3, C2 & \#B3, C3 & \#B3, C4 & \#B3, C5 \\
    \hline
    3D Mean & 76.8  & 77.4  & \textbf{78.1} & 76.6  & 77.6  & 77.6  & 78.2  & 77.1  & \textbf{78.1} & 76.8 \\
    2D Mean & 85.0 & 85.0  & 84.8  & 84.7  & 84.9  & 84.5  & 84.6  & \textbf{85.1} & 84.8  & 84.5 \\
    Mean  & 79.7  & 80.1  & \textbf{80.5} & 79.5  & 80.2  & 80.0    & 80.5  & 79.9  & \textbf{80.5} & 79.5 \\
    \hline\hline
    \end{tabular}%
    }
  \label{tab: fuse module}%
\end{table*}%

\begin{table}[t]
  \centering
  \caption{Results of the modal-specific and universal task prompts with varying shapes. The best results are highlighted in bold.}
   \resizebox{1.0\columnwidth}{!}{
    \begin{tabular}{c|cccccc|>{\centering\arraybackslash}p{1.5cm}>{\centering\arraybackslash}p{1.5cm}>{\centering\arraybackslash}p{1.5cm}}
    \hline\hline
    \multirow{2}[4]{*}{Prompt Shape} & \multicolumn{6}{c|}{Modal Prompts ($l$)}      & \multicolumn{3}{c}{Universal Task Prompt ($K\times4\times6\times6$)} \\
\cmidrule{2-10}          & 256   & 320   & 384   & 512   & 1024  & 2048  & 50    & 100   & 200 \\
    \hline
    3D Mean & 77.4  & 76.9  & 77.4  & \textbf{78.1}  & 77.9  & 77.9  & 77.5  & \textbf{78.1}  & 77.6 \\
    2D Mean & \textbf{85.1}  & 84.8  & 84.7  & 84.8  & 84.8  & 84.5  & 84.6  & 84.8  & 84.6 \\
    Mean  & 80.2  & 79.7  & 80.0  & \textbf{80.5}  & 80.4  & 80.2  & 80.0  & \textbf{80.5}  & 80.0 \\
    \hline\hline
    \end{tabular}%
    }
  \label{tab: prompt shape}%
\end{table}%

\begin{table}[t]
  \centering
  \caption{Performance of 2D models (\textit{i.e.}, 2D UniMiSS and 2D Backbone) and 3D models (\textit{i.e.}, 3D Backbone and MedUniSeg) on the SIIM dataset.}
  \resizebox{1.0\columnwidth}{!}{
    \begin{tabular}{c|cccc}
    \hline\hline
 Method & \#Param. (M) & GPU Mem. (M) & Inference Time (s/Image) & Dice \\
    \hline
    UniMiSS  & 26.47 & 690   & 0.298 & 54.8 \\
    2D Backbone  & 10.71 & 614   & 0.093 & 55.6 \\
    3D Backbone  & 31.17 & 880   & 0.171 & 55.7 \\
    MedUniSeg & 31.24 & 938   & 0.173 & 59.8 \\
    \hline\hline
    \end{tabular}%
    }
  \label{tab: inference}%
\end{table}%

\subsection{Comparing to Other Pre-trained Models}
To verify the transfer ability of our MedUniSeg, we compared it with recent advanced models, including both self-supervised and supervised models. The self-supervised models include single-modal pre-trained models such as MG \cite{zhou2021models}, GVSL \cite{he2023geometric}, DeSD \cite{ye2022desd}, SMIT \cite{jiang2022self}, Swin UNETR \cite{tang2022self}, VoCo \cite{wu2024voco}, BT \cite{kang2023benchmarking}, PCRLv2 (CheXpert) \cite{zhou2023unified}, and multi-modal pre-trained models like MedKLIP \cite{wu2023medklip}, and UniMiSS \cite{xie2022unimiss}. The supervised models include Universal Model \cite{liu2023clip}, Hermes \cite{gao2024training}, DoDNet \cite{zhang2021dodnet}, CCQ \cite{liu2023ccq}, and UniSeg \cite{ye2023uniseg}. Moreover, we introduced two nnUNet models trained from scratch, one with a 3D backbone (representing MedUniSeg without pre-trained weights) and the other with a 2D backbone, highlighting the improvements gained from pre-training. The 2D backbone was derived from the 3D one by replacing 3D modules with 2D counterparts. 
For employing 3D pre-trained models on 2D tasks, similar to upstream training, we treated 2D data as pseudo 3D data. 
Models such as SMIT, Swin UNETR, VoCo, and Universal models$\dag$, implemented based on the Swin Transformer \cite{liu2021swin}, cannot be directly applied to 2D tasks, since the depth length of the Swin Transformer must be greater than 1. Furthermore, these models are also unsuitable for the $48\times192\times192$ patch size used on the BTCV and VS datasets due to depth requirements. Consequently, we adopted a patch size of $64\times192\times192$ for these models. For UniMiSS, we followed the official strategy of forming two models to address 2D and 3D tasks, respectively. All results are averaged over three runs to ensure robustness.

The results in Table \ref{tab: fine-tuning} reveal several findings: 
(1) Our MedUniSeg significantly outperforms its baseline, the 3D backbone, across all downstream datasets, regardless of whether 20\%, 50\%, and 100\% training data is used. This indicates that universal learning enables MedUniSeg to acquire high-quality representations, boosting downstream task performance.
(2) Compared to other pre-trained models, MedUniSeg exhibits the best performance across all datasets, except for the COVID-19-20 dataset with 20\% training data, where it secures the second-best performance. This performance advantage stems from MedUniSeg's robust representation capability, allowing it to outperform most self-supervised and supervised models.
(3) When compared to UniSeg, which was pre-trained on three modalities, MedUniSeg, pre-trained on nine modalities, shows consistent improvement across all datasets. This underscores the benefits of learning from a broader range of modalities and richer data.
(4) Although the parameters of the 3D backbone are approximately three times larger than those of the 2D backbone, it achieves comparable performance. Nevertheless, using a 3D backbone to address both 2D and 3D tasks remains superior to employing a Transformer backbone. Further discussion is provided in Section \ref{inference}.

\subsection{Ablation Studies}
We evaluated six variants of MedUniSeg, including UniSeg, Multiple Prompts, Universal Prompts, Fixed Prompts, Bottleneck Prompts, and MedUniSeg-T.
Fig. \ref{fig: variants} illustrates the structures of MedUniSeg and its variants. The differences between MedUniSeg and these variants are as follows: 
UniSeg is regarded as MedUniSeg without modal priors.
Multiple Prompts employs modal- and task-specific prompts to generate corresponding priors.
Universal Prompts uses universal modal and task prompts to generate modal and task priors, respectively.
Fixed Prompts functions as MedUniSeg with fixed prompts.
Bottleneck Prompts incorporates both modal and task priors at the end of the encoding process.
MedUniSeg-T includes task priors at the end of the decoding process.

The results, presented in Table \ref{tab: ablation}, demonstrate the superior performance of our MedUniSeg in 3D mean Dice, 2D mean Dice, and overall mean Dice.
More importantly, we validated the motivations behind this study by comparing MedUniSeg with these variants. 
Compared to UniSeg, our findings confirm the effectiveness of the proposed modal prior.
Comparisons with Multiple Prompts and Universal Prompts reveal that combining modal-specific prompts with a universal task prompt is the most effective strategy for capturing correlations and providing priors for modalities and tasks.
Additionally, the comparison with Fixed Prompts highlights the advantage of using learnable prompts over fixed alternatives. 
Further comparisons with Bottleneck Prompts and MedUniSeg-T confirm the optimal positions for integrating modal and task priors.
In summary, MedUniSeg exemplifies the optimal configuration of modal and task priors, validated in terms of both their introduction and positioning.

\subsection{Block and Channel Numbers of FUSE Module}
Our FUSE module consists of multiple convolutional blocks designed to reduce input feature channels from $C$ to $C/m$ in the first block, ultimately outputting $N$ channels, each corresponding to a specific task.
Here, $C$ is the sum of the channels from the universal task prompt and the sample-specific features.
We conducted a detailed assessment of the module's design, focusing on the number of blocks ($\#B$) and the channel ($C/m$). With $C$ fixed, we examined the impact of varying the reduction ratio $m$.
The results in Table \ref{tab: fuse module} indicate that with $m$ fixed at four, MedUniSeg achieves the highest mean Dice score when using three blocks. Conversely, with $\#B$ fixed at three, the optimal mean Dice score is obtained by setting $m$ to four. Thus, the combination of three blocks and a $C/4$ channel configuration offers the most effective task-specific priors, leading to superior generalization performance.

\subsection{Shapes of Modal-specific Prompts and Universal Task Prompt}
We conducted experiments to vary the length of the modal-specific prompt ($l$) and the channel number of the universal task prompt ($K$), with the results summarized in Table \ref{tab: prompt shape}. For $l$, we tested six variants, gradually increasing its value from 256 to 2048. Among these, setting $l$ to 512 yielded the highest mean Dice score and 3D mean Dice. Similarly, for $K$, we evaluated values of 50, 100, and 200, determining that $K = 100$ is optimal. Consequently, in our MedUniSeg, we set $l$ to 512 and $K$ to 100.

\subsection{Visualization of Segmentation Results}
\subsubsection{Upstream dataset}
We visualized segmentation results obtained from UKAN, UMamba, nnUNet, Universal Model, Hermes, DoDNet, CCQ, UniSeg, and MedUniSeg across seven datasets, as illustrated in Fig. \ref{fig: upstream visualization}. The visualizations demonstrate that MedUniSeg’s segmentation results closely align with the ground truths (GTs), effectively mitigating under-segmentation (see the first row of Fig. \ref{fig: upstream visualization}) and over-segmentation (see the third row of Fig. \ref{fig: upstream visualization}). Moreover, compared to UniSeg (our previous work), MedUniSeg consistently delivers more accurate results across all images, highlighting the advancements achieved in this version.
\subsubsection{Downstream datasets}
We visualized the segmentation results of several models, including Swin UNETR, UniMiSS, DeSD, Universal Model, Universal Model$\dag$, Hermes, DoDNet, CCQ, UniSeg, and MedUniSeg, across six downstream datasets. A representative sample from each dataset was presented in Fig. \ref{fig: downstream visualization}. The visualizations clearly demonstrate that MedUniSeg consistently outperforms competing methods in terms of accuracy across five modalities and both 2D and 3D segmentation tasks. For instance, in images from the SIIM and SegPC datasets, MedUniSeg not only provides the most complete segmentation but also minimizes over-segmentation compared to other methods.

%=====================================================================================================================================
%============================================================Discussion===============================================================
%=====================================================================================================================================
\section{Discussion}
\subsection{Visualization of Task-specific Prior}
To investigate the features learned by the universal task prompt, we visualized the task-specific priors using t-SNE. These task-specific priors were derived from all training and test data. Due to imbalanced sample sizes across tasks, we randomly selected 1,000 samples from each task for the t-SNE visualization. For tasks with fewer than 1,000 samples, we employed a resampling strategy to augment the data to this threshold. The resulting visualizations are presented in Fig. \ref{fig: visual task prompt}. The t-SNE visualization reveals that the distributions of different tasks are well-clustered and exhibit clear classification boundaries. This indicates that the task-specific priors learned through the self-learn universal task prompt can effectively distinguish and describe the unique characteristics of each task, thereby minimizing prompt confusion within the model. For instance, despite tasks like Liver, Kidney, HepaV, Pancreas, Colon, Lung, and Spleen segmentation sharing similar input images, their task priors display significant distributional differences.

% The limitation of t-SNE is that it is unsuitable for describing the correlations between tasks, therefore, we adopted the Pearson correlation to analyze these correlations. The task-specific priors of each task are averaged to get a representative task-specific prior for calculation. To simplify, we used 0.6 as a threshold to screen out results with absolute values greater than the threshold and visualize them with a chord diagram.
% (2) by the chord diagram, we found that correlations between tasks are complex and hard-craft with many counterintuitive correlations. For example, there is a similarity of 0.69 between the task-specific prompts from the VerSE20 (CT vertebrae segmentation) and Prostate (MRI prostate segmentation). Moreover, we also observed an intriguing phenomenon that the strong correlations (similarity$>$0.6) only exist within tasks with identical dimensions. 

\subsection{Correlation between Upstream and Downstream Learning}
A limitation of self-supervised learning is the challenge of evaluating the transferability of a pre-trained model using upstream metrics, such as loss value. Importantly, lower loss values do not necessarily correlate with better downstream performance. In the context of supervised pre-training, we investigated whether upstream performance metrics could serve as reliable predictors for downstream performance. To this end, we calculated the correlation between upstream and downstream performance. Specifically, for each universal model, we recorded the mean Dice score across 17 upstream datasets and the Dice scores on six downstream datasets with 100\% training data. We then computed Pearson correlation coefficients for each dataset. The Pearson correlation coefficients between upstream performance and the BTCV, COVID-19-20, VS, SIIM, IDRID, and SegPC datasets were 0.75, 0.74, 0.87, 0.57, 0.64, and 0.61, respectively, indicating positive correlations in most cases. Therefore, we conclude that the transferability of a supervised pre-trained model can generally be assessed by its upstream performance.

\subsection{Resource Requirements for Inference} \label{inference}
In this study, we utilized a 3D UNet architecture to handle both 3D and 2D segmentation tasks, treating 2D data as pseudo-3D data with a depth of one. However, this approach inherently leads to inefficiencies for 2D tasks, as parameters assigned to the depth dimension have minimal impact. We recorded the number of parameters, GPU memory usage, inference times per image, and Dice scores for UniMiSS, 2D backbone, 3D backbone, and MedUniSeg, as summarized in Table \ref{tab: inference}. All models were tested on an RTX 3090 with a batch size of 1 and patch size of $512\times512$ using the nnUNet framework. The results indicate that although MedUniSeg requires approximately twice the inference time and 1.5 times the GPU memory compared to the 2D version, it achieves a 4.2\% improvement in Dice scores. More importantly, when compared to the Transformer-based model UniMiSS, which also supports both 2D and 3D input, MedUniSeg outperforms it in both Dice scores and inference times.

In summary, MedUniSeg offers a superior solution for both 2D and 3D segmentation, achieving better performance and lower inference times compared to UniMiSS.

%=====================================================================================================================================
%============================================================Conclusion===============================================================
%=====================================================================================================================================

\section{Conclusion}
In this paper, we present MedUniSeg, a prompt-driven universal model specifically designed for 2D and 3D medical image segmentation across diverse targets, modalities, and domains. Our approach integrates modal-specific prompts and a universal task prompt to effectively characterize both the modalities and tasks. Utilizing these prompts, we develop the MMap and FUSE modules to generate modal- and task-specific priors, which are strategically incorporated at the start and end of the encoding process, respectively.
We evaluate MedUniSeg on a large-scale multi-modal segmentation upstream dataset and six downstream segmentation datasets. The results demonstrate its superior performance in both universal learning and transfer learning.
For tasks that exhibit suboptimal performance during the initial multi-task joint training, we freeze MedUniSeg and introduce new LoRA modules, deconvolutional layers, and segmentation heads to re-learn these tasks, resulting in an enhanced version called MedUniSeg*. This strategy consistently improves task performance compared to the original MedUniSeg.
In the future, we plan to integrate MedUniSeg to address medical image classification and detection tasks, further enhancing its universality.

% or


% % use section* for acknowledgment
% \ifCLASSOPTIONcompsoc
%   % The Computer Society usually uses the plural form
%   \section*{Acknowledgments}
% \else
%   % regular IEEE prefers the singular form
% \fi


\ifCLASSOPTIONcaptionsoff
  \newpage
\fi



%\bibliographystyle{IEEEtran}
\bibliographystyle{IEEEtran}
\bibliography{mybib}

% \end{thebibliography}


\begin{IEEEbiography}[{\includegraphics[width=1in,height=1.25in,clip,keepaspectratio]{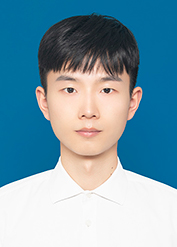}}]{Yiwen Ye}
received his B.E. degree in computer science and technology in 2021 from Hebei University of Technology, Tianjin, China. He is currently working toward the Ph.D. degree at the School of Computer Science and Engineering, Northwestern Polytechnical University (NPU), Xi’an, China. His research interests include universal segmentation models and representation learning.
Biography text here.
\end{IEEEbiography}
\begin{IEEEbiography}[{\includegraphics[width=1in,height=1.25in,clip,keepaspectratio]{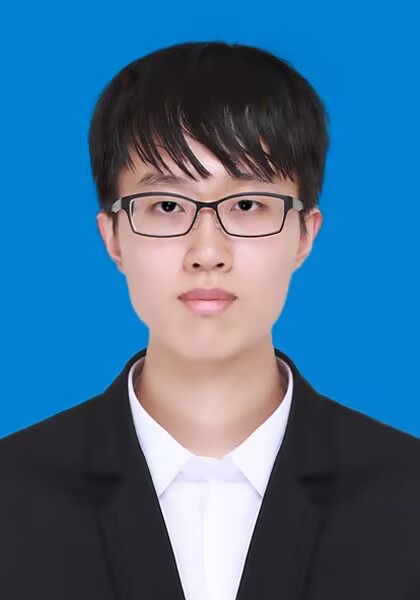}}]{Ziyang Chen}
received his B.E. degree in biological technology in 2021 from Northwestern Polytechnical University, Xi’an, China. He is currently working toward the Ph.D. degree at the School of Computer Science and Engineering, Northwestern Polytechnical University, Xi’an, China. His research interests include domain adaptation and domain generalization.
\end{IEEEbiography}
\begin{IEEEbiography}[{\includegraphics[width=1in,height=1.25in,clip,keepaspectratio]{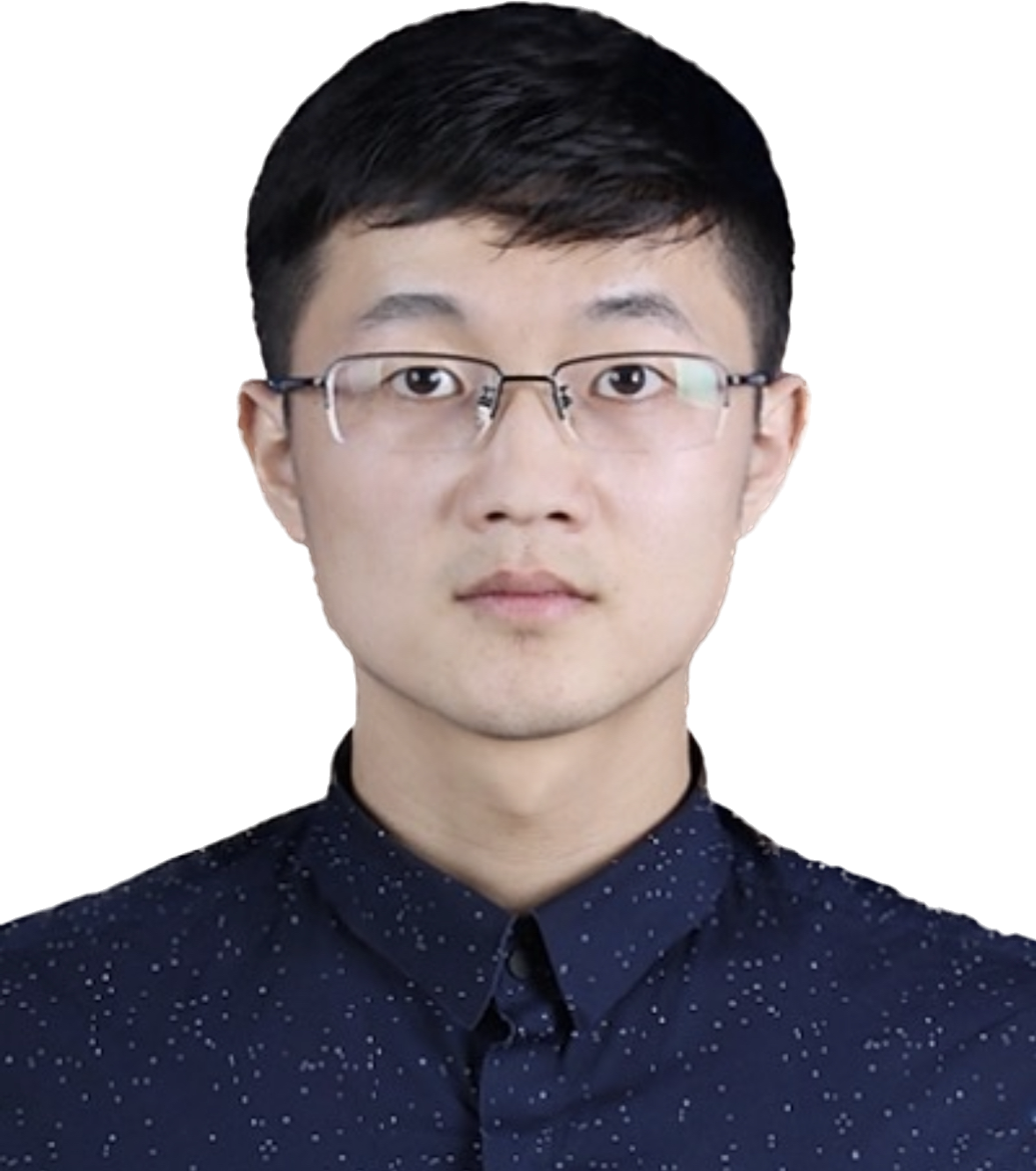}}]{Jianpeng Zhang}
received the PhD degree in Computer Science and Technology from Northwestern Polytechnical University, China, in 2022. His research interests mainly focus on deep learning technologies for intelligent medical image analysis, especially medical vision-language learning, self-supervised learning, partial label learning, and weakly supervised learning.
\end{IEEEbiography}
\begin{IEEEbiography}[{\includegraphics[width=1in,height=1.25in,clip,keepaspectratio]{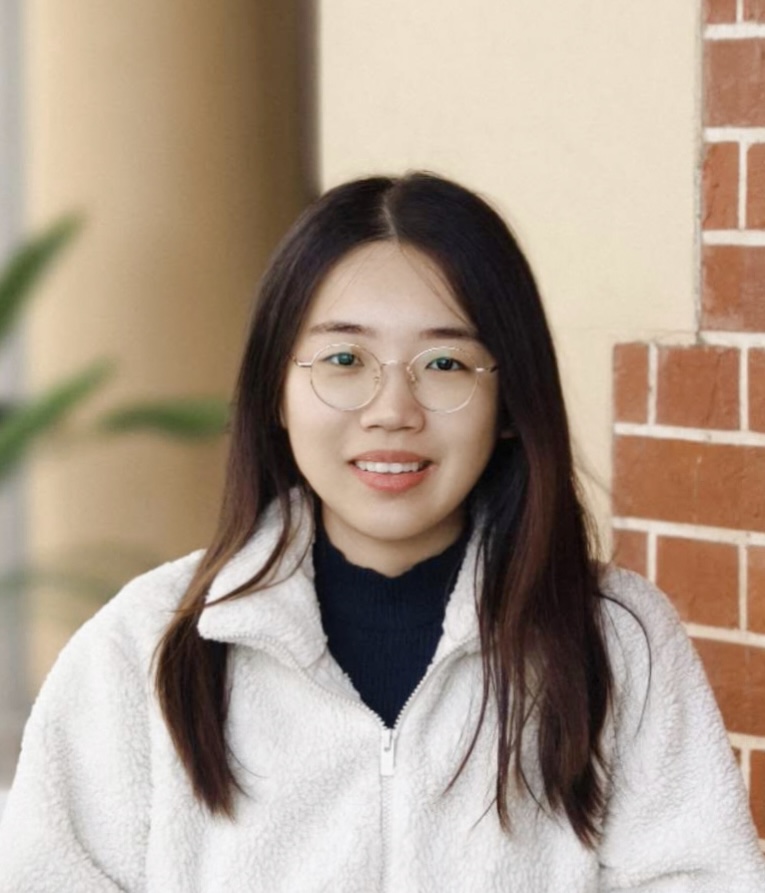}}]{Yutong Xie}
received her B.E. degree in 2016 and her Ph.D. in 2021 from Northwestern Polytechnical University (NPU), Xi’an, China. She is currently a Senior Research Fellow at the University of Adelaide (UoA) and a member of the Australian Institute for Machine Learning (AIML). Her research primarily focuses on computer vision and data analytics within the healthcare sector, aiming to develop intelligent solutions to assist healthcare professionals in anatomical structure segmentation, disease diagnosis, and therapy.
\end{IEEEbiography}
\begin{IEEEbiography}[{\includegraphics[width=1in,height=1.25in,clip,keepaspectratio]{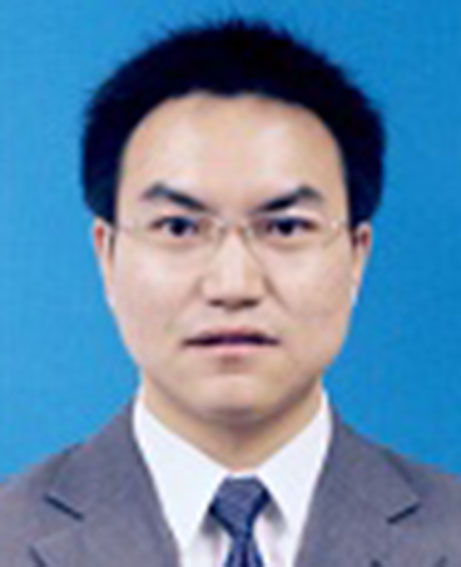}}]{Yong Xia}(S'05-M'08) received his B.E., M.E., and Ph.D. degrees in computer science and technology from Northwestern Polytechnical University (NPU), Xi’an, China, in 2001, 2004, and 2007, respectively. He is currently a Professor at the School of Computer Science and Engineering, NPU. His research interests include medical image analysis, computer-aided diagnosis, pattern recognition, machine learning, and data mining.
\end{IEEEbiography}


% % insert where needed to balance the two columns on the last page with
% %\newpage



%\vfill


\end{document}